\pdfoutput=1
\documentclass[11pt]{article}

\usepackage[]{acl}

\usepackage{times}
\usepackage{latexsym}
\usepackage{booktabs}
\usepackage{graphicx}
\usepackage{tikz}
\usepackage{bbding}
\usepackage{multirow}
\usepackage{amsmath}

\usepackage[inline]{enumitem}

\usepackage[T1]{fontenc}
\usepackage[utf8]{inputenc}

\usepackage{microtype}

\usepackage{inconsolata}

\definecolor{forestgreen}{rgb}{0.13, 0.55, 0.13}

\usepackage{colortbl}

\title{Robust Text Classification: Analyzing Prototype-Based Networks}

\author{Zhivar Sourati$^{1,2}$, Darshan Deshpande$^{1,2}$, Filip Ilievski$^{1,3}$, \\ \textbf{Kiril Gashteovski$^{4,5}$} \and \textbf{Sascha Saralajew$^{4}$}\\ $^{1}$Information Sciences Institute, University of Southern California, Marina del Rey, CA, USA \\ $^{2}$Department of Computer Science, University of Southern California, Los Angeles, CA, USA
\\ $^{3}$Department of Computer Science, Vrije Universiteit Amsterdam, The Netherlands \\ $^{4}$NEC Laboratories Europe, Heidelberg, Germany \\ $^{5}$CAIR, Ss. Cyril and Methodius University, Skopje, North Macedonia}

\begin{document}
\maketitle
\begin{abstract}
Downstream applications often require text classification models to be accurate and robust. While the accuracy of state-of-the-art Language Models (LMs) approximates human performance, they often exhibit a drop in performance on real-world noisy data. This lack of robustness can be concerning, as even small perturbations in text, irrelevant to the target task, can cause classifiers to incorrectly change their predictions. A potential solution can be the family of Prototype-Based Networks (PBNs) that classifies examples based on their similarity to prototypical examples of a class (prototypes) and has been shown to be robust to noise for computer vision tasks. In this paper, we study whether the robustness properties of PBNs transfer to text classification tasks under both targeted and static adversarial attack settings.
Our results show that PBNs, as a mere architectural variation of vanilla LMs, offer more robustness compared to vanilla LMs under both targeted and static settings. We showcase how PBNs' interpretability can help us understand PBNs' robustness properties. Finally, our ablation studies reveal the sensitivity of PBNs' robustness to the strictness of clustering and the number of prototypes in the training phase, as tighter clustering and a low number of prototypes result in less robust PBNs. 
\end{abstract}

% Introduction section

\section{Introduction}
\label{sec:introduction}

Language models (LMs) are widely used in various NLP tasks and exhibit exceptional performance \citep{chowdhery2022palm,zoph2022stmoe}. In light of the need for real-world applications of these models, the requirements for robustness and interpretability have become urgent for both foundational Large Language Models (LLMs) and fine-tuned LMs. More fundamentally, robustness and interpretability are essential components of developing trustworthy technology that can be adopted by experts in any domain~\citep{wagstaff2012machine,Slack2022TalkToModelEM}. However, LMs have limited interpretability by design \citep{zhao2023explainability,gholizadeh2021model}, which cannot be fully mitigated by posthoc explainability techniques~\citep{10.1145/3529755}. Moreover, LMs lack robustness when exposed to text perturbations, noisy data, or distribution shifts 
\citep{jin2020bert,moradi2021evaluating}. Reportedly, even LLMs lack robustness when faced with out-of-distribution data and noisy inputs \citep{wang2023robustness}, a finding that is supported by the empirical findings of this paper, too. 

\begin{figure*}[!ht]
    \centering
    \includegraphics[width=0.85\textwidth]{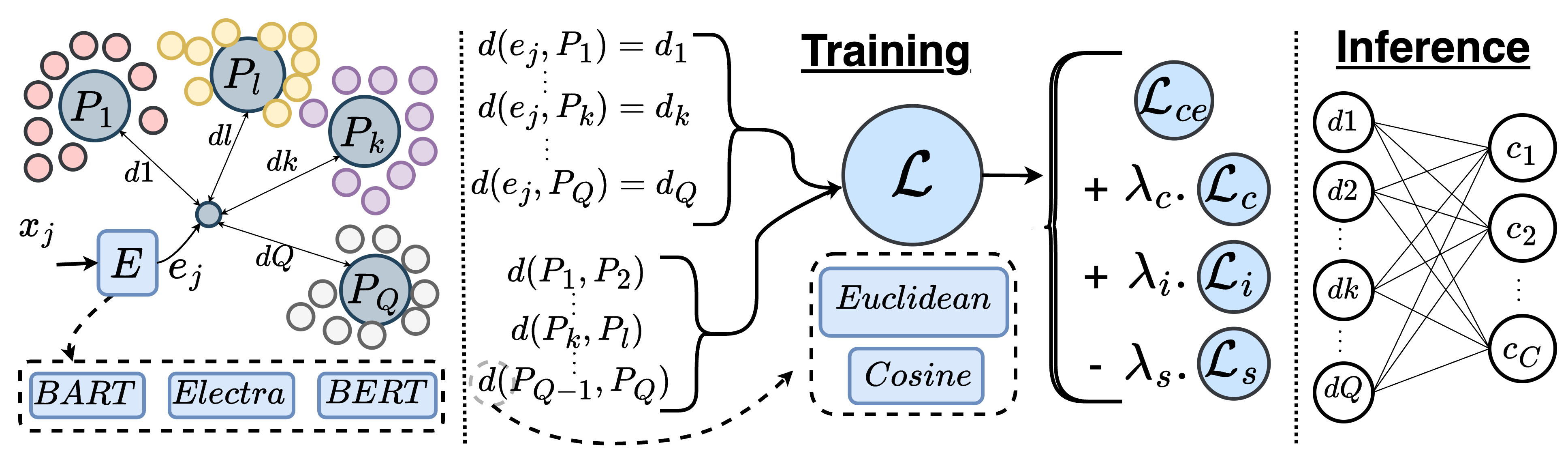}
    \caption{Classification by a PBN\@. The model computes distances between the new point and prototypes, $d(e_j, P_k)$, and distances within prototypes, $d(P_k, P_l)$, for both inference and training. During training, the model minimizes the loss term, $\mathcal{L}$, consisting of $\mathcal{L}_{ce}$, $\lambda_c \mathcal{L}_{c}$, $\lambda_i \mathcal{L}_{i}$, $\lambda_s \mathcal{L}_{s}$, controlling the importance of accuracy, clustering, interpretability, and separation of prototypes, based on all the computed distances; during inference, distances between the new point and prototypes are used for classification by a fully connected layer.}
    \label{fig:architecture}
\end{figure*}

On this ground, NLP research has increasingly focused on benchmarks, methods, and studies that emphasize robustness and interpretability~\citep[e.g., ][]{zhou2020rica,jang2022becel,Liu2021ExplainaBoardAE}. This has also been accompanied by the surge of focus on models that are inherently and architecturally interpretable and robust \citep[e.g.,][]{koh2020concept,papernot2018deep,keane2019case}.
An example of such models is the family of Prototype-Based Networks (PBNs) that is designed for robustness and interpretability~\citep{li_cbr_prototypes_2018}. PBNs are based on the theory of categorization in cognitive science~\citep{ROSCH1973328}, where categorization is governed by the graded degree of possessing prototypical features of different categories, with some members being more central (\textit{prototypical}) than others. Consider, for example, classifying different types of birds. Then, pelican classification can be done through their prototypical tall necks and similarity to a prototypical pelican~\citep{Nauta_2021}. Computationally, this idea is implemented by finding prototypical points or examples in the shared embedding space of data points and using the distance between prototypes and data points to accomplish the classification task. As noted by \citet{linzen2020can}, human-like classification approaches also rely on distances to prototypical examples, which PBNs leverage to achieve robustness levels similar to humans. The use of prototypes allows PBNs to compare input points to prototypical examples, making them resilient to small perturbations in the input (e.g., changes in phrases or words) without losing the overall semantic meaning. This is because the classification task in PBNs utilizes multiple prototypes and their corresponding distances, which represent different semantic aspects of the data, thereby enhancing robustness against noise and adversarial attacks~\citep{yang2018robust}. Furthermore, prototypes implicitly maximize the margin between classes as shown in certain adversarially robust architectures~\citep{voravcek2022provably}.
% , though how this extends to latent space when inputs are attacked remains an open area for further research.

PBNs have been popular in Computer Vision (CV) tasks, including image classification~\citep{angelov2020towards} and novel class detection~\citep{hase2019interpretable}. Inspired by PBNs in CV, NLP researchers have also developed PBN models for text classification, in particular, for sentiment classification~\citep{protocnn,Ming_2019,hong2021interpretable}, few-shot relation extraction~\citep{han-etal-2021-few-shot-re,meng2023rapl}, and propaganda detection~\citep{prototex}. Yet, while competitive performance and interpretability of PBNs have been studied in both NLP~\citep{prototex,Hase2020} and CV~\citep{gu2019hierarchical,vanaken2022patient}, their robustness advantages over vanilla models have only been investigated in CV~\citep{yang2018robust,Saralajew2020,Voracek2022}.

In this study, \textit{we investigate whether the robustness properties of PBNs transfer to NLP classification tasks}. In particular, our contributions are:
(1) We adopt a modular and comprehensive approach to evaluate PBNs' robustness properties against various well-known black-box adversarial attacks under both targeted and static adversarial settings;
(2) We conduct a comprehensive analysis of the sensitivity of PBNs' robustness with respect to different hyperparameters. 
    % (3) We perform \textbf{extensive experiments} on four benchmarks to compare the robustness of PBNs to vanilla models, 

Our experiments show that PBNs improve the robustness of language models in text classification tasks by effectively handling realistic perturbations under both targeted and static adversarial settings. We note that the robustness boost that adversarial augmented training brings to LMs with access to additional pieces of relevant data is higher than the boost caused by PBNs' architecture. Nevertheless, considering that the robustness boost in PBNs is only caused by their architecture without any additional resources, and this architecture is interpretable by design, the merits of such models can contribute to the field. Finally, benefiting from inherent interpretability, we showcase how PBN interpretability properties help to explain PBNs' robust behavior. We release our code to support future research.\footnote{\url{https://github.com/zhpinkman/robust-prototype-learning}}

\section{Prototype-Based Networks}
\label{sec:pbn}

PBNs classify data points based on their similarity to a set of \textit{prototypes} learned during training. These prototypes summarize prominent semantic patterns of the dataset through two mechanisms: (1) prototypes are defined in the same embedding space as input examples, which makes them interpretable by leveraging input examples in their proximity; and (2) prototypes are designed to cluster semantically similar training examples, which makes them representative of the prominent patterns embedded in the data and input examples.
The PBN's decisions, based on quantifiable similarity to prototypes, are robust as noise and perturbations are better reflected in the computed similarity to familiar prototypical patterns~\citep{hong2020interpretable}. Additionally, prototypes can provide insight during inference by helping users explain the model's behavior on input examples through the prototypes utilized for the model's prediction~\citep{prototex}. 
% \textcolor{red}{this part can be dropped in favor of needing space in other important parts: Prototypes being in the same embedding space as input examples allows them to be represented as either the training examples \citep{prototex} or parts of training examples, such as key phrases \citep{protocnn} or key sequences \citep{Ming_2019,hong2021interpretable} extracted from training examples. These prototypes can be associated with semantic patterns of particular classes from their initialization or be trained freely and subsequently associated with the prominent semantic patterns of the whole dataset.}

% \vspace*{0.1cm}
\paragraph{Inference.} Classification in PBNs is done via a fully connected layer applied on the measured distances between embedded data points and prototypes. As shown in \autoref{fig:architecture}, given a set of data points $x_j, j \in \{1,\dots, N\}$ with labels $y_j \in \{1,\dots, C\}$, and $Q$ prototypes, PBNs first encode examples with a backbone $E$, resulting in the embedding $e_j = E(x_j)$. Next, PBNs compute the distances between prototypes and $e_j$ using the function $d$. These distances get fed into a fully connected layer to compute class-wise logits, incorporating the similarities to each prototype. Applying a softmax on top, the final outputs are $\hat{y}_{c}(x_j)$: probability that $x_j$ belongs to class $c \in \{1,\dots, C\}$.

% \vspace*{0.1cm}
\paragraph{Training.} The model is trained using objectives that simultaneously tweak the backbone parameters and the (randomly initialized) prototypes, thus promoting high performance and meaningful prototypes. To compute a total loss term $\mathcal{L}$, PBNs use the computed distances within prototypes $d(P_k, P_l)_{k \neq l}$, distances between all $Q$ prototypes and $N$ training examples given by $d(e_j, P_k)_{j\in\{1,\dots, N\}; k\in\{1,\dots, Q\}}$, and the computed probabilities $\hat{y}_c$. The prototypes and the weights in the backbone are adjusted according to $\mathcal{L}$. The total loss $\mathcal{L}$ consists of different inner loss terms that ensure high accuracy, clustering, interpretability, and low redundancy among prototypes; i.\,e., the classification loss $\mathcal{L}_{ce}$, the clustering loss $\mathcal{L}_{c}$ \citep{li_cbr_prototypes_2018}, the interpretability loss $\mathcal{L}_{i}$ \citep{li_cbr_prototypes_2018}, and separation loss $\mathcal{L}_{s}$ \citep{hong2020interpretable}:

\vspace{-1em}
\begin{equation}
    \mathcal{L} = \mathcal{L}_{ce} + \lambda_c \mathcal{L}_{c} + \lambda_i \mathcal{L}_{i} - \lambda_s \mathcal{L}_{s},
\label{eq:main_loss}
\end{equation}

\noindent where $\lambda_c, \lambda_i, \lambda_s \geq 0$ are regularization factors to adjust the contribution of the auxiliary loss terms.

\textit{Classification loss $\mathcal{L}_{ce}$} is defined as the cross-entropy loss between predicted and true labels:

\vspace{-1em}
\begin{equation}
    \mathcal{L}_{ce} = -\sum_{j=1}^N\log(\hat{y}_{y_{j}}(x_j)).
\end{equation}

\textit{Clustering loss} $\mathcal{L}_{c}$ ensures that the training examples close to each prototype form a cluster of similar examples. In practice, $\mathcal{L}_{c}$ keeps all the training examples as close as possible to at least one prototype and minimizes the distance between training examples and their closest prototypes:

\vspace{-1em}
\begin{equation}
    \mathcal{L}_{c} = \frac{1}{N} \sum_{j = 1}^{N} \min_{k \in \{1,\dots, Q\}} d(P_k, e_j).
\label{eq:clustering_loss}
\end{equation}

\textit{Interpretability loss $\mathcal{L}_{i}$} ensures that the prototypes are interpretable by minimizing the distance to their closest training sample: 

\vspace{-1em}
\begin{equation}
    \mathcal{L}_{i} = \frac{1}{Q} \sum_{k = 1}^{Q} \min_{j\in\{1,\dots, N\}} d(P_k, e_j).
\label{eq:interpretability_loss}
\end{equation}

\noindent Keeping the prototypes close to training samples allows PBNs to represent a prototype by its closest training samples that are domain-independent and enable analysis by task experts.

\textit{Separation loss} $\mathcal{L}_{s}$ maximizes the inter-prototype distance to reduce the probability of redundant prototypes:

\vspace{-1em}
\begin{equation}
    \mathcal{L}_{s} = \frac{2}{Q(Q-1)} \sum_{k, l \in \{1,\dots, Q\}; k\neq l; } d(P_k, P_l).
\label{eq:separation_loss}
\end{equation}

\section{Robustness Evaluation}
\label{sec:robustness-evaluation}

\begin{table}
\centering
\small
\resizebox{\linewidth}{!}{%
\begin{tabular}{l|l}
    \toprule
    \textbf{Original text} & \textbf{Perturbed text} \\ 
    \midrule
    A g\textcolor{teal}{e}ntle \textcolor{teal}{breeze} rustled the l\textcolor{teal}{e}aves. & A g\textcolor{red}{è}ntle \textcolor{red}{wind} rustled the l\textcolor{red}{E}aves.\\
    \midrule
     \textcolor{teal}{r}es\textcolor{teal}{c}ue \textcolor{teal}{Engineer} Company & \textcolor{red}{R}es\textcolor{red}{©}ue \textcolor{red}{operation} Company \\
    \midrule
    embarrassing\textcolor{teal}{l}y fo\textcolor{teal}{o}lish & embarrassing\textcolor{red}{1}y fo\textcolor{red}{0}lish \\
    % \midrule
    % \textcolor{teal}{movie}'s special pl\textcolor{teal}{ac}e & \textcolor{red}{film}'s special pl\textcolor{red}{ca}e \\
    \bottomrule
\end{tabular}}
\caption{Examples of adversarial perturbations, with the perturbed tokens highlighted.}
\label{tab:perturbation-examples}
\vspace{-0.2em}
\end{table}

We assess PBNs' robustness against adversarial perturbations of original input text that are intended to preserve the text's original meaning. The perturbations change the classification of the target model upon confronting these perturbed examples from the correct behavior to an incorrect one in an effective and efficient way \citep{10.1145/1014052.1014066,kurakin2017adversarial_a,kurakin2017adversarial_b,li2023efficiently}. \textbf{Strategies} for finding these perturbations vary \citep{zhang2020adversarial}: perturbations can be focused on different granularities, i.e., \textit{character-level}, \textit{word-level}, or \textit{sentence-level}; their generation can be done in different ways, e.g., \textit{replacing}, \textit{inserting}, \textit{deleting}, \textit{swapping} tokens; they can have different searching strategies for their manipulations, such as \textit{context-aware} or \textit{isolated} approaches; and also various salient token identification strategies to maximize their adversarial effect.

Orthogonally, adversarial perturbations are divided into \textbf{targeted} and \textbf{static}. In the targeted setting, the attacker aims to mislead a specific model toward incorrect predictions \citep{si2021better}. However, in the static setting, adversarial examples are crafted without targeting any particular model. Instead, the same perturbations are generated by attacking external models that the attacker has access to, referred to as source models. These successful perturbations are then applied to a different model, the target model, which is the one being evaluated for robustness. The performance of the target model against these pre-collected perturbations is used to assess its robustness \citep{adv-glue}.

Additionally, adversarial attacks can be categorized as \textbf{white-box} or \textbf{black-box} \citep{zhang2020adversarial}. A white-box attack occurs when the attacker has knowledge of the target model's architecture and parameters. This access allows for a more precise generation of adversarial examples explicitly tailored to exploit vulnerabilities in the target model. In contrast, a black-box attack is performed without access to the target model’s internal details; instead, the attacker generates adversarial examples by querying the model or using outputs from other accessible models to infer information about the target model’s behavior. The robustness of the target model is then evaluated based on how effectively it withstands generated adversarial examples. In this study, we specifically focus on \textbf{black-box} attacks under both \textbf{targeted} and \textbf{static} settings.

With numerous adversarial perturbation strategies in the literature \citep{zhang2020adversarial,wang2022measure}, each with unique advantages (e.g., effectiveness vs. efficiency), we use a wide range of existing perturbation strategies in this study. These cover the aforementioned granularities, generation strategies, searching strategies, and salient token identification strategies under both targeted and static settings. See examples of adversarial perturbations covered in our study in \autoref{tab:perturbation-examples}.

\vspace{-0.3em}
\section{Experimental Setup}
\label{sec:experimental-setup}

% First, we present the classification datasets we select to adversarially attack the classifiers trained on them (see \autoref{subsec:datasets}). In \autoref{subsec:metrics-perturbations}, we present the details about the adversarial attacks, perturbations, and the metrics used to report the robustness of models against adversarial attacks. In \autoref{subsec:pbns-hyperparameters}, we iterate over the hyperparameters we used to train the PBNs, and finally, we present the baselines in \autoref{subsec:baselines}.
\vspace{-0.3em}
\subsection{Datasets} 
\label{subsec:datasets}

PBNs classify instances based on their similarity to prototypes learned during training that summarize prominent semantic patterns in a dataset. Thus, with more classes, we might need more prototypes to govern the more complex system between instances and prototypes~\citep{yang2018robust}. To study the interplay between the number of classes and robustness, we employ three datasets: 
(1) \textit{IMDB reviews} \citep{maas-EtAl:2011:ACL-HLT2011}: a binary sentiment classification dataset; 
(2) \textit{AG\_NEWS} \citep{GulliAG}: a collection of news articles that can be associated with four categories; 
(3) \textit{DBPedia}:\footnote{\url{https://bit.ly/3RgX41H}} a dataset with taxonomic, hierarchical categories for Wikipedia articles \citep{lehmann2015dbpedia}, with nine classes. We use these three datasets to study the robustness of PBNs under both targeted and static adversarial settings. As an additional source of static adversarial perturbations, we adopt the SST\nobreakdashes-2 binary classification split from the existing \textit{Adversarial GLUE (AdvGLUE)} dataset~\citep{adv-glue}, consisting of perturbed examples of different granularities, filtered both automatically and by human evaluation for more effectiveness. For statistics of the datasets and their perturbations, see \autoref{appendix:dataset-details}. 

\subsection{Perturbations} 
\label{subsec:metrics-perturbations}

As mentioned before, we will focus on black-box adversarial perturbations under both targeted and static settings. However, there are various strategies to produce perturbations given original input examples. In the following, we will elaborate on the strategies we used as well as how they were utilized in the targeted and static settings.

\paragraph{Attacking strategies.}
To produce perturbations given original inputs, we selected five well-established adversarial attack strategies: BAE \citep{garg-ramakrishnan-2020-bae}, TextFooler \citep{jin2020bert}, TextBugger \citep{li2018textbugger}, DeepWordBug \citep{gao2018black}, and PWWS \citep{ren-etal-2019-generating}.\footnote{We also employed paraphrased-based perturbations \citep{lei2019discrete}, generated by GPT3.5 \citep{chatgpt}. However, both our baselines and PBNs were robust to these perturbations, and we include them in the Appendix in \autoref{tab:robustness-of-pbns-paraphrased-based-perturbations}.} As mentioned in \autoref{sec:robustness-evaluation}, these attacks cover a wide range of granularities (e.g., character-based in DeepWordBug and word-based in PWWS), generation strategies (e.g., word substitution in PWWS and TextFooler and deletion in TextBugger), searching strategies (e.g., context-aware in BAE and isolated synonym-based in TextFooler), and salient token identification strategies (e.g., finding the important sentences first and then words in TextBugger and finding the important words to change in BAE). For details of how each attack strategy works to produce perturbations given original input examples, refer to \autoref{subsec:perturbation-details}.

\paragraph{Targeted perturbations.}
In this setting, the adversarial attacks are directly conducted against PBNs and vanilla LMs trained on original datasets. For each attack strategy, we aim for 800 successful perturbations and report the robustness of PBNs against adversarial attacks by Attack Success Rate (ASR; \citealp{wu2021performance}) and Average Percentage of Words Perturbed (APWP; \citealp{yoo2020searching}) to reach the observed ASR\@. Successful perturbations are those that change the prediction of a target model already fine-tuned on that dataset from the correct prediction to the wrong prediction.

\paragraph{Static perturbations.}
In this setting, the adversarial attacks are conducted on external models: BERT \citep{devlin2018bert}, RoBERTa \citep{liu2019roberta}, and DistilBERT \citep{sanh2019distilbert}, which are trained on the original datasets, and a compilation of the successful perturbations on those models is used to assess the robustness of PBNs against the studied adversarial attacks by their accuracy on the perturbations, similar to the study by \citet{adv-glue}. To obtain the perturbations, each model is fine-tuned on each dataset, and 800 successful perturbations for each attack strategy are obtained. We focus on examples whose perturbations are predicted incorrectly by all three models to maximize the generalizability of this static set of perturbations to a wider range of unseen target models. In principle, the perturbations for each model are different, yielding three variations per original example for a dataset-perturbation pair. For instance, focusing on DBPedia and BAE attack strategy, after 800 successful perturbations for each of the three target models, the perturbations of 347 original examples could change all models' predictions, resulting in a total of 1401 (3\,$\times$\,347) perturbations compiled for BAE attack strategy and DBPedia dataset. 

\subsection{PBNs' Hyperparameters}
\label{subsec:pbns-hyperparameters}

% \vspace*{0.1cm}
\paragraph{Backbone ($E$).} 
Prototype alignment and training are highly dependent on the quality of the latent space created by the backbone encoder $E$, which in turn affects the performance, robustness, and interpretability of PBNs. We consolidate previous methods for text classification using PBNs \citep{protocnn,prototex,Ming_2019,hong2020interpretable} and consider three backbone architectures: BERT \citep{devlin2018bert}, BART encoder \citep{lewis2019bart}, and Electra \citep{clark2020electra}. Based on our empirical evidence, fine-tuning all the layers of the backbone was causing the PBNs' training not to converge. Hence, we freeze all the layers of the backbones except for the last layer when training. 

% \vspace*{0.1cm}
\paragraph{Distance function ($d$).} 
The pairwise distance calculation quantifies how closely the prototypes are aligned with the training examples (\autoref{fig:architecture}). In recent work, Euclidean distance has been shown to be better than Cosine distance for similarity calculation \citep{vanaken2022patient,NIPS2017_cb8da676} as it helps to align prototypes closer to the training examples in the encoder's latent space. However, with some utilizing Cosine distance \citep{this_looks_like_that_2019} while others prioritizing Euclidean distance \citep{mettes2019hyperspherical}, and the two having incomparable experimental setups, conclusive arguments about the superiority of one over the other cannot be justified, and the choice of distance function is usually treated as a hyperparameter. Accordingly, we hypothesize that the impact of $d$ will be significant in our study of robustness, and hence, we consider both Cosine and Euclidean distance functions when training PBNs.

% \vspace*{0.1cm}
\paragraph{Number of prototypes ($Q$).} 
Number of prototypes in PBNs is a key factor for mapping difficult data distributions~\citep{yang2018robust,sourati2023robust}. Hence, to cover a wide range, we consider five values for $Q = \{2, 4, 8, 16, 64\}$.

% \vspace*{0.1cm}
\paragraph{Objective functions ($\mathcal{L}$).} 
Given the partly complementary goals of loss terms, we investigate the effect of interpretability, clustering, and separation loss on PBNs' robustness, keeping the accuracy constraint ($\mathcal{L}_{ce}$) intact. To do so, we consider three values, $\{0, 0.9, 10\}$ for $\lambda_i$, $\lambda_c$, and $\lambda_s$. $0$ value represents the condition where the corresponding loss function is not being utilized in the training process. $0.9$ value was empirically found to offer good accuracy, clustering, and interpretability, across datasets and was also motivated by prior works \citep{prototex}. $10$ value was chosen as an upper bound dominating the corresponding loss objective (e.g., interpretability) in the training process.

\begin{table*}[ht]
\aboverulesep=0ex % Solution part 1 of 3
\belowrulesep=0ex
\centering
\small
\setlength\tabcolsep{4.5pt}
\resizebox{\linewidth}{!}{%
\begin{tabular}{@{\extracolsep{\fill}}rrrrrr|rrrrr|rrrrr}
\multicolumn{12}{c}{\normalsize Targeted Attacks; Attack Success Rate (ASR \%) reported}  \\
\toprule
& \multicolumn{5}{c|}{AG\_News} & \multicolumn{5}{c|}{DBPedia} & \multicolumn{5}{c}{IMDB} \\
 & BAE & DWB & PWWS & TB & TF & BAE & DWB & PWWS & TB & TF & BAE & DWB & PWWS & TB & TF \\
\midrule
BART & 14.8 & 53.2 & 53.6 & 31.8 & 76.5 & 18.9 & 28.3 & 43.1 & 21.1 & 71.9 & 74.1 & 74.7 & 99.3 & 78.5 & 100.0 \\
+ PBN & \textbf{11.1} & \textbf{32.3} & \textbf{41.3} & \textbf{23.1} & \textbf{62.2} & \textbf{15.2} & \textbf{14.7} & \textbf{28.7} & \textbf{12.6} & \textbf{45.5} & \textbf{36.1} & \textbf{41.0} &\textbf{75.9} & \textbf{41.3} & \textbf{73.1} \\
\midrule
BERT & 17.0 & 78.0 & 69.8 & 45.7 & 88.8 & 13.9 & 24.8 & 31.6 & 22.0 & 61.3 & 82.5 & 79.7 & 99.9 & 83.9 & 99.9 \\
+ PBN & \textbf{7.7} & \textbf{42.6} & \textbf{47.0} & \textbf{30.4} & \textbf{70.5} & \textbf{9.8} & \textbf{17.3} & \textbf{21.6} & \textbf{13.0} & \textbf{41.0} & \textbf{42.8} & \textbf{41.0} & \textbf{79.7} & \textbf{57.7} & \textbf{79.8} \\
\midrule
ELEC. & 24.8 & 89.5 & 69.1 & 87.8 & 87.9 & 14.5 & 42.8 & 45.6 & 42.3 & 75.3 & 52.5 & 49.2 & 95.3 & 67.8 & 99.3 \\
+ PBN & \textbf{14.0} & \textbf{34.9} & \textbf{42.9} & \textbf{51.8} & \textbf{70.2} & \textbf{7.8} & \textbf{11.5} & \textbf{17.8} & \textbf{19.1} & \textbf{35.6} & \textbf{28.9} & \textbf{27.4} & \textbf{66.6} & \textbf{36.8} & \textbf{78.0} \\
\bottomrule
\end{tabular}}
\newline
\vspace{0.3em}
\resizebox{\linewidth}{!}{%
\begin{tabular}{@{\extracolsep{\fill}}rrrrrr|rrrrr|rrrrr|r}
\multicolumn{9}{c}{\large Static Attacks; Accuracy (\%) reported}  \\
\toprule
& \multicolumn{5}{c|}{AG\_News} & \multicolumn{5}{c|}{DBPedia} & \multicolumn{5}{c|}{IMDB} & SST2\\
 & BAE & DWB & PWWS & TB & TF & BAE & DWB & PWWS & TB & TF & BAE & DWB & PWWS & TB & TF & GLUE\\
\midrule
BART & 53.2 & 76.7 & 83.2 & 77.5 & 85.8 & 55.5 & 68.6 & 58.4 & 72.5 & 71.3 & 74.1 & 80.5 & 83.6 & 85.8 & 87.6 & \underline{29.8} \\
+ PBN & \underline{57.6} & \textbf{80.6} & \underline{84.8} & \textbf{79.2} & \underline{88.8} & \underline{65.0} & \underline{71.6} & \underline{65.7} & \underline{78.4} & \underline{74.8} & \underline{80.4} & \underline{81.3} & \underline{86.3} & \underline{89.3} & \underline{90.4} & \textbf{50.4} \\
+ Aug. & \textbf{71.7} & \underline{78.4} & \textbf{85.5} & \underline{77.6} & \textbf{90.1} & \textbf{84.0} & \textbf{79.6} & \textbf{89.7} & \textbf{88.8} & \textbf{94.0} & \textbf{85.7} & \textbf{86.7} & \textbf{92.9} & \textbf{89.9} & \textbf{96.5} & - \\
\midrule
BERT & 47.8 & 64.0 & 75.9 & 69.4 & 80.7 & 62.3 & 61.4 & 75.4 & 78.4 & 82.0 & 75.1 & 77.1 & 85.0 & 83.4 & 85.9 & \underline{42.0} \\
+ PBN & \underline{52.9} & \underline{70.4} & \textbf{78.5} & \textbf{73.8} & \underline{84.3} & \underline{66.9} & \underline{66.6} & \underline{80.3} & \underline{82.0} & \underline{85.8} & \underline{77.6} & \textbf{79.1} & \underline{85.3} & \underline{85.0} & \underline{86.5} & \textbf{51.1} \\
+ Aug. & \textbf{58.3} & \textbf{71.6} & \underline{78.3} & \underline{71.2} & \textbf{85.4} & \textbf{75.5} & \textbf{70.9} & \textbf{84.1} & \textbf{90.5} & \textbf{91.0} & \textbf{83.2} & \underline{77.6} & \textbf{91.7} & \textbf{90.8} & \textbf{89.2} & - \\
\midrule
ELEC. & 50.4 & \underline{65.0} & \underline{73.5} & \underline{63.9} & 77.8 & \underline{79.7} & 66.9 & \underline{80.9} & 81.4 & 84.4 & \underline{89.7} & 90.3 & \underline{94.6} & 94.5 & 95.6 & \underline{44.3} \\
+ PBN & \textbf{64.6} & \textbf{74.1} & \textbf{85.1} & \textbf{77.2} & \textbf{89.0} & 78.7 & \underline{69.8} & 79.3 & \underline{82.5} & \underline{85.8} & \textbf{90.0} & \underline{90.8} & \underline{94.6} & \textbf{95.5} & \textbf{96.3} & \textbf{65.6} \\
+ Aug. & \underline{55.0} & 59.5 & 71.7 & 61.6 & \underline{79.5} & \textbf{86.2} & \textbf{73.8} & \textbf{88.1} & \textbf{84.5} & \textbf{92.8} & 89.4 & \textbf{93.7} & \textbf{95.3} & \underline{94.9} & \underline{95.8} & - \\
\midrule
GPT4o & 57.1 & 73.3 & 73.0 & 76.5 & 79.9 & 66.0 & 63.4 & 61.0 & 69.0 & 44.0 & 87.0 & 89.5 & 91.2 & 93.7 & 94.2 & 59.8 \\
Llama3 & 57.6 & 56.4 & 55.0 & 65.9 & 62.8 & 44.0 & 53.7 & 37.8 & 45.0 & 44.4 & 82.0 & 86.0 & 93.2 & 89.0 & 91.5 & 56.0 \\
\bottomrule
\end{tabular}}
\caption{
Comparison of PBNs and vanilla LMs (+ vanilla LMs with adversarial augmented training under static attack setting) under both targeted and static adversarial attack perturbations, using the best hyperparameters for PBNs, on IMBD, AG\_News, DBPedia (+ SST-2 from AdvGLUE under static attack setting) datasets, under BAE, DeepWordBug (DWB), PWWS, TextBugger (TB), TextFooler (TF). The highest accuracy and lowest ASR showing the superior model for each architecture is \textbf{boldfaced}, and the second best model is \underline{underlined} for static attacks.}
\label{tab:new-table-results-both}
\vspace{-0.3em}
\end{table*}

\vspace{-0.5em}
\subsection{Baselines}
\label{subsec:baselines}

Since PBNs are architectural enhancements of vanilla LMs using learned prototypes for classification instead of a traditional softmax layer used in vanilla LMs, \textbf{vanilla LMs} employed as PBNs' backbones serve as a baseline for comparing the robustness of PBNs. We also employ \textbf{adversarial augmented training} \citep{goyal2023survey} on top of the vanilla LMs as another baseline. Note that the same layers frozen for PBNs' training are also frozen for the baselines. As we need additional data for adversarial augmented training, we use this baseline under static perturbations, where the set of perturbations has already been compiled beforehand. Finally, we compare PBNs with two LLMs, namely, GPT4o \citep{gpt_4o_2024} and Llama3 \citep{llama3modelcard}. Although we note that LLMs are more appropriate choices for generic chat and text generation due to their decoder-only architecture, and fine-tuned LMs might still be superior to LLMs when it comes to task-oriented performance \citep{chang2024survey}, we include this comparison for comprehensiveness. The superiority of smaller fine-tuned LMs is confirmed in our experiments, too (see \autoref{tab:performance-of-models-on-original-datasets} in Appendix), where smaller fine-tuned models beat LLMs on the original datasets.

% Results

\section{Results}
\label{sec:results}

\vspace{-0.6em}
\subsection{Robustness of PBNs}
\label{subsec:robustness-of-pbns}

The robustness report of PBNs under both targeted and static adversarial attacks under different experimental setups (i.e., datasets, backbones, and attack strategies), using the best hyperparameters is presented in \autoref{tab:new-table-results-both}.~\footnote{To ensure that the perturbations do not significantly affect the meaning of original texts, we compared the two versions in terms of semantic similarity using OpenAI text-embedding-ada-002 across all datasets and attack types, which yielded a high similarity: 0.97 ($SD = 0.01$).}~\footnote{Our results showed that adversarial perturbations from TextFooler and PWWS were more effective than others.} Best hyperparameters were chosen among the permutation of all hyperparameters presented in \autoref{subsec:pbns-hyperparameters} to yield the highest robustness (lowest ASR or highest accuracy). Under the targeted setting, our results showed that PBNs are more robust than vanilla LMs (having lower ASR) regardless of the utilized backbone, dataset, or attacking strategy. We saw similar trends analyzing the robustness of PBNs compared to vanilla LMs, averaging over all PBN hyperparameters (find the details in \autoref{tab:robustness-of-pbn-on-average}). Focusing on the APWP metric, the PBNs' robustness was greater than vanilla LMs (having higher APWP) in $71.0\%$ of the conditions, and this superiority dropped to $31.0\%$ of the conditions when averaging over all the hyperparameters (find the details in \autoref{tab:robustness-of-pbns-apwp}), which suggested that PBNs' robustness is sensitive to hyperparameters involved in training. 

We observed similar trends under static adversarial attacks, where the PBNs' robustness was higher than vanilla LMs (having higher accuracy under attack) in the majority of the conditions ($93.7\%$ of all variations of experimental setups and hyperparameters). Furthermore, in each experimental condition (dataset and attack strategy), there was always a PBN that outperformed larger LLMs, i.e., GPT4o \citep{gpt_4o_2024} and Llama 3 \citep{llama3modelcard}, which have significantly more parameters but lack the interpretability that PBNs inherently offer.
% Upon closer investigation, the only conditions in which PBNs were close to but not more robust than vanilla LMs were the PWWS attacks done on the ELECTRA backbone fine-tuned on the IMDB and DBPedia datasets. 
Vanilla LMs with adversarial augmented training demonstrated greater robustness than PBNs in $71.2\%$ of the conditions. This highlighted the more effective role of additional data in adversarial augmented training compared to PBNs' robust architecture, which makes PBNs a preferable choice when efficiency is prioritized \citep{goodfellow2014explaining}. Analyzing PBNs' robustness under the static adversarial setting averaging over all PBNs' hyperparameters, our results showed that in only $31.2\%$ of the conditions, PBNs have greater robustness compared to vanilla LMs (find the details in \autoref{tab:robustness-of-pbn-on-average}), which similar to observations on APWP, suggested that PBNs' robustness is sensitive to hyperparameters involved in the training.

To sum up, we observed that PBNs consistently and over different metrics were more robust compared to vanilla LMs and LLMs, using the best hyperparameters without sacrificing performance on the original unperturbed samples (find performance on original datasets in \autoref{tab:performance-of-models-on-original-datasets}). We believe that the observed robust behavior is due to the design of the PBN architecture. Standard neural networks for text classification distinguish classes by drawing hyperplanes between samples of different classes that are prone to noise \citep{yang2018robust}, especially when dealing with several classes. Instead, PBNs are inherently more robust since they perform classification based on the similarity of data points to prototypes, acting as class centroids. Finally, we observed that the robustness superiority of PBNs compared to vanilla LMs diminished when averaging over all possible hyperparameters, indicating that the robustness of PBNs is sensitive to hyperparameter choices. We investigated this sensitivity further in \autoref{subsec:pbns-robustness-sensitivity-hyperparameters}.

\vspace{-0.4em}
\subsection{Sensitivity to Hyperparameters}
\label{subsec:pbns-robustness-sensitivity-hyperparameters}
\vspace{-0.3em}
We studied the sensitivity of PBNs' robustness to the hyperparameters involved in training, covering values discussed in \autoref{subsec:pbns-hyperparameters}. Focusing on each hyperparameter, the value for the other ones was selected to yield the best performance so that, overall, we could better depict the sensitivity and limiting effect of the hyperparameter of interest. We did not observe any sensitivity from PBNs with respect to the backbone, interpretability term ($\lambda_i$; see \autoref{app:subsec:effect-of-interpretability-robustneess}), separation term ($\lambda_s$; see \autoref{app:subsec:effect-of-separation-robustneess}), and the distance function ($d$; see \autoref{app:subsec:effect-of-distance-function}). 

\begin{figure}
\centering
\includegraphics[width=0.8\columnwidth]{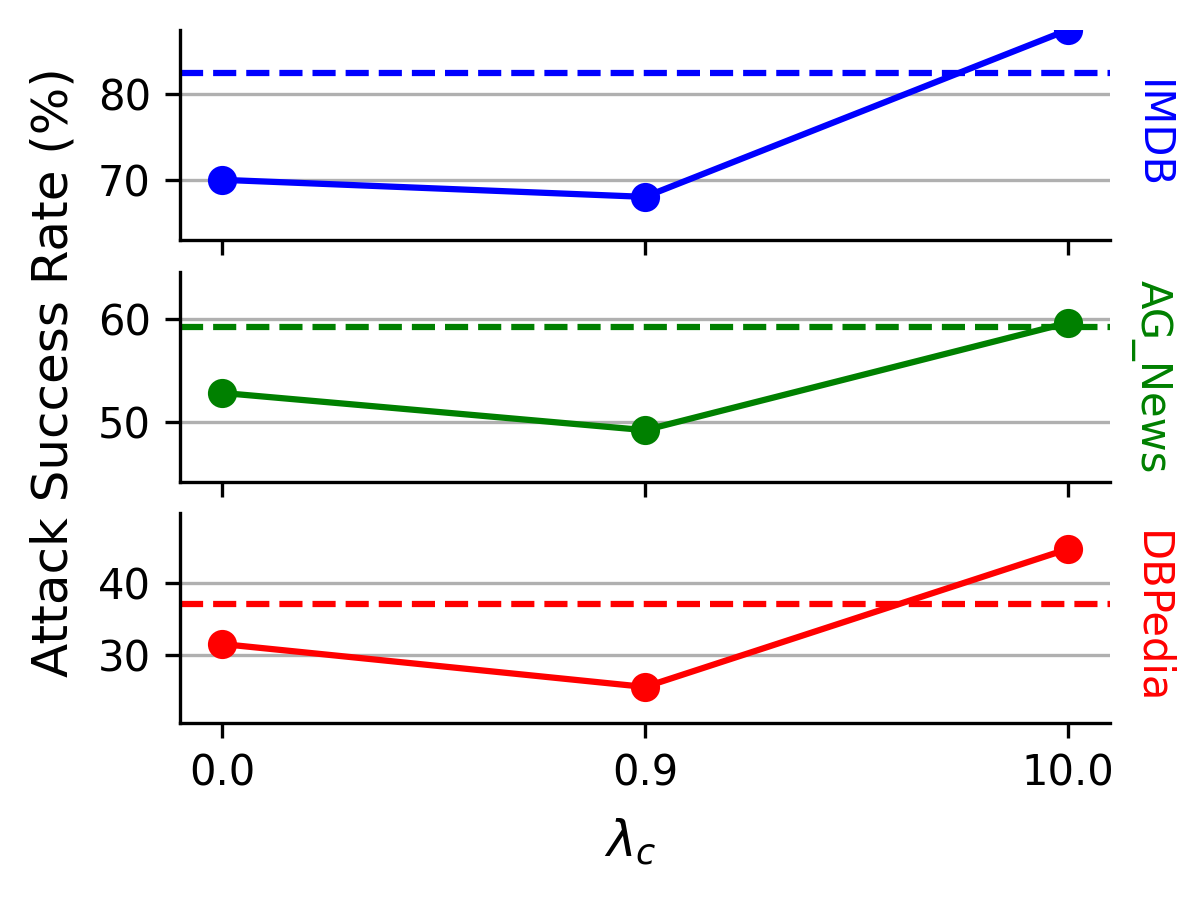}
    \caption{Attack Success Rate (ASR \%) of PBNs with different $\lambda_c$ values adjusting the importance of clustering in the trained PBNs, with other hyperparameters set to their best values, and averaged across other possible variables (e.g., backbone and attack type). The dashed line represents the ASR for the non-PBN model.}
    \label{fig:asr_effect_of_pc}
    \vspace{-0.4em}
\end{figure}

However, as presented in \autoref{fig:asr_effect_of_pc}, we observed that higher values of $\lambda_c$, promoting tighter clustering of input examples around prototypes, hinder PBNs' robustness. Clustering loss is a regularization term that encourages samples to be close to prototypes in the embedding space, further enhancing interpretability but potentially reducing accuracy by narrowing the diversity in embedding space, which is a common phenomenon in loss terms of competing goals. The mean and standard deviation over (transformed) distances between prototypes and samples can be used to describe the spread of embedded data points around prototypes. These values are $(-0.24\pm\!1.7)\times10^{-7}$ with $\lambda_c = 0.9$, and $(-0.18\pm\!1.5)\times10^{-6}$ with $\lambda_c = 10$, showing less diverse prototypes indicated by smaller measured distances caused by stronger clustering. 

\begin{figure}
    \centering
    \includegraphics[width=0.8\columnwidth]{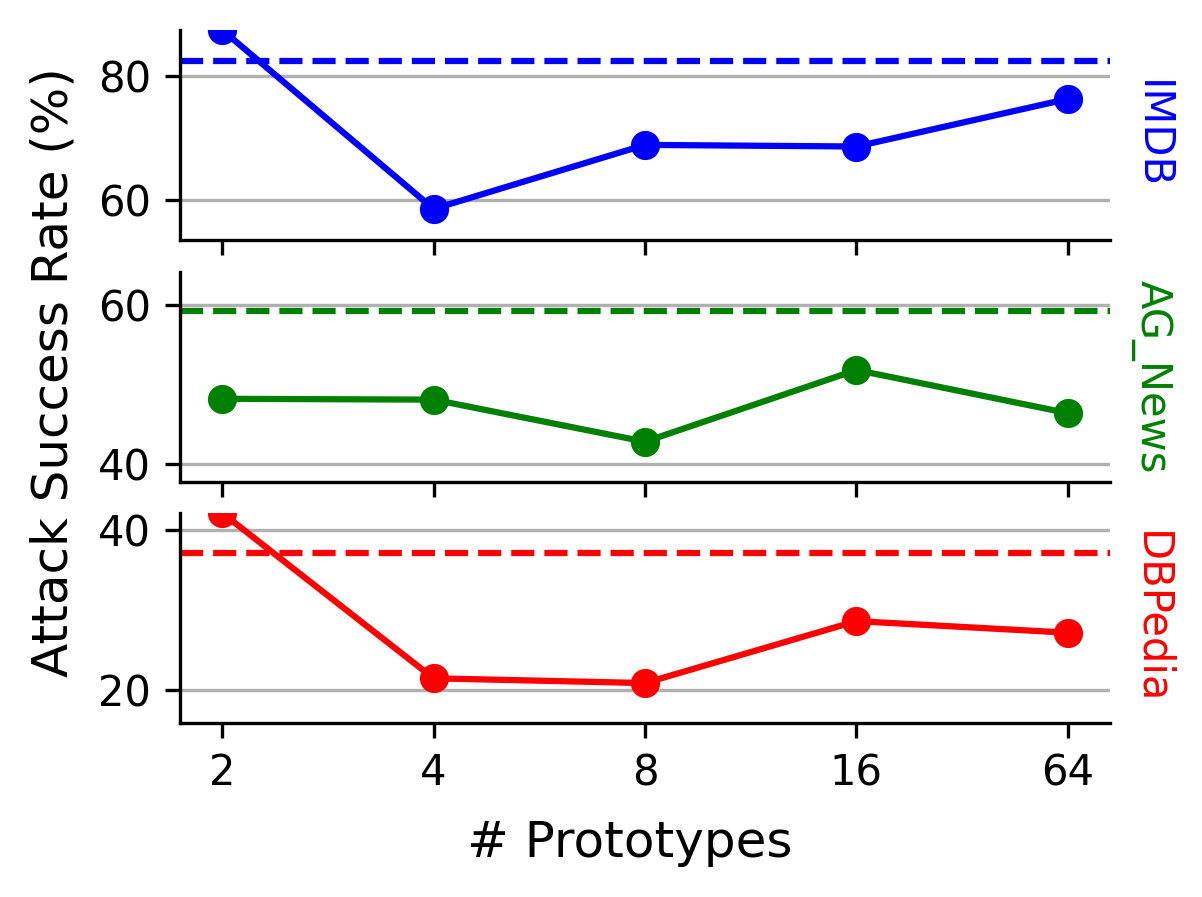}
    \caption{Attack Success Rate (ASR \%) of PBNs with different numbers of prototypes, with other hyperparameters set to their best values, and averaged across other possible variables (e.g., backbone and attack type). dashed line represents the ASR for the non-PBN model.}
    \label{fig:effect-of-num-prototypes}
    \vspace{-0.3em}
\end{figure}

Additionally, as depicted in \autoref{fig:effect-of-num-prototypes}, we observed poor robustness from PBNs when the number of prototypes is as low as two, which is intuitive as a low number of prototypes also means a lower number of semantic patterns learned, which constraints the PBNs' abilities to distinguish between different classes. Noting that more prototypes add to the complexity and size of the network as a whole, the observed stable trend of the robustness with the higher number of prototypes ($> 2$) suggests that as long as the number of prototypes is not too low, PBNs with lower number of prototypes can be preferred. This corroborates with the studies performed by \citet{yang2018robust}. Finally, note that the same analysis using other metrics (e.g., APWP) and under static adversarial setting (using accuracy as the studied metric) depicted the same trend and can be found in \autoref{app:subsec:effect-of-clustering-robustneess} and \autoref{app:subsec:effect-of-num-prototypes-on-robustness}.

\begin{table}
\aboverulesep=0ex % Solution part 1 of 3
\belowrulesep=0ex
\centering
\setlength\tabcolsep{4.5pt}
\resizebox{1.0\columnwidth}{!}{%
\begin{tabular}{p{0.10\columnwidth}|p{0.85\columnwidth}|p{0.15\columnwidth}}
\toprule
Proto. & Representative Training Examples & Label \\
\midrule
$P_0$ & \textbf{Handly's Lessee v. Anthony (1820)}: Determined Indiana-Kentucky boundary. & UnitWork \\
& \textbf{Rasul v. Bush (2004)}: Decided jurisdiction over Guantanamo detainees. & UnitWork \\
\midrule
$P_1$ & \textbf{Marine Corps Air Station Futenma}: U.S. Marine Corps base, Ginowan, Okinawa; regional military hub. & Place \\
& \textbf{Özdere}: Turkish coastal resort town in İzmir Province, popular among tourists. & Place \\
\midrule
$P_2$ & \textbf{Yevgeni Viktorovich Balyaikin}: Russian footballer for FC Tom Tomsk. & Agent\\
& \textbf{Gigi Morasco}: Fictional character on ABC's One Life to Live. & Agent \\
\bottomrule
\end{tabular}}
\caption{Examples of prototypes, their closest training examples, alongside their label derived from their closest training examples, extracted from a PBN with 16 prototypes and a BART backbone on DBPedia. Note that the presented training examples are the summarization of their longer version for easier interpretation. }
\label{tab:prototypes}
\end{table}

\begin{table*}
\aboverulesep=0ex 
\belowrulesep=0ex
\centering
\small
\setlength\tabcolsep{4.5pt}
\resizebox{\linewidth}{!}{%
\begin{tabular}{p{0.75\linewidth}|p{0.11\linewidth}|p{0.11\linewidth}|p{0.04\linewidth}|p{0.04\linewidth}}
\toprule
Text & Activ. Proto.s & Proto.s Labels & Pred. & Label \\
\midrule
\textbf{Roman Catholic Diocese of Barra}: Diocese in Barra, Feira de Santana province, Brazil. & $P_1, P_{14}$ & Place, Place & Place & Place\\
\textbf{Roman Catholic Bishop of Barra}: Episcopal seat in Barra, Feira de Santana province, Brazil. & $P_1, P_{14}$ & Place, Place & Place & Place \\
\midrule
\textbf{Inta Ezergailis}: Latvian American professor emerita at Cornell University. & $P_2, P_8$ & Agent, Agent & Agent & Agent\\
\textbf{Inta Ezergailis}: Latvian American poet and scholar at Cornell University. & $P_2, P_7$ & Agent, Work & Agent & Agent\\
\midrule
\textbf{Saint Eigrad}: 6th-century Precongregational North Wales saint and Patron Saint of Llaneigrad. & $P_2, P_8$ & Agent, Agent & Agent & Agent \\
\textit{\textbf{St Eigrad}: 6th-century Precongregational street of North Wales and Patron Saint of Llaneigrad.} & $P_1, P_{14}$ & Place, Place & Place & Agent \\ 
\bottomrule
\end{tabular}}
\caption{Examples of the original (top) and adversarially perturbed (bottom) examples of DBPedia using TextFooler, classified by a PBN, alongside the top-2 activated prototypes by the PBN's fully connected layer and their associated labels. Incorrectly predicted examples are in \textit{italic}.}
\label{tab:prototypes-robustness}
\vspace{-0.3em}
\end{table*}

\vspace{-0.4em}
\subsection{PBNs’ Interpretability w.r.t. Robustness}
\label{subsec:pbns-robustness-interpretability}

PBNs are interpretable by design \citep{this_looks_like_that_2019}, and we can understand their behavior through the distance of input examples to prototypes and the importance of these distances, extracted by the last fully connected layer of PBNs transforming vector of distances to log probabilities for classes. While proving interpretability in the traditional sense is beyond the scope of this paper (refer to \citealp{prototex,hoffmann2021looks,davoodi2023interpretability,ragno2022prototype} for more in-depth analysis of interpretability of PBNs), we showcase how prototypes in PBNs can be interpretable and utilized for robustness analysis under adversarial attacks. 

Examples of learned prototypes that can be represented by their closest training input examples are shown in \autoref{tab:prototypes}. These input examples help the user identify the semantic features that the prototypes are associated with, which by our observations in our case, were mostly driven by the class label of the closest training examples. We can also benefit from interpretable properties of PBNs to better understand their robustness properties, regardless of the success of perturbations. \autoref{tab:prototypes-robustness} illustrates predictions of a PBN on three original and perturbed examples from the DBPedia dataset, alongside the top-2 prototypes that were utilized by the PBN's fully connected layer for prediction and prototypes' associated label (by their closest training examples). In the first two examples, PBN correctly classifies both the original and perturbed examples, and from the top-2 prototypes, we observe that this is due to unchanged prototypes utilized in prediction. However, in the last example, the model incorrectly classifies an example that is associated with an Agent as a Place. Interestingly, this incorrect behavior can be explained by the change in the top-2 activated prototypes, where they are changing from Agent-associated to Place-associated prototypes because of the misspelling of "saint" with "street." Thus, the use of prototypes not only enhances our understanding of the model's decision-making process but also unveils how minor perturbations influence the model's predictions.

% Related Work
\vspace{-0.5em}
\section{Related Work}
\label{sec:related-work}
\vspace{-0.5em}
\paragraph{Robustness evaluation.}
% \paragraph{Robustness evaluation.}
Robustness in NLP is defined as models' ability to perform well under noisy \citep{ebrahimi-etal-2018-hotflip} and out-of-distribution data \citep{hendrycks-etal-2020-pretrained}. With the wide adoption of NLP models in different domains and their near-human performance on various benchmarks \citep{wang2019glue,sarlin2020superglue}, concerns have shifted towards models' performance facing noisy data \citep{adv-glue,wang2022recode}. Studies have designed novel and effective adversarial attacks \citep{jin2020bert,zhang2020adversarial}, defense mechanisms \citep{goyal2023survey,liu2020adversarial}, and evaluations to better understand the robustness properties of NLP models \citep{adv-glue,morris2020reevaluating}. These evaluations are also being extended to LLMs, as they similarly lack robustness \citep{wang2023robustness,shi2023large}.
While prior work has studied LMs' robustness, 
to our knowledge, PBNs' robustness properties have not been explored yet. Our study bridges this gap.

% \vspace*{0.1cm}
\paragraph{Prototype-based networks.}
PBNs are widely used in CV \citep{this_looks_like_that_2019,hase2019interpretable,Kim_2021_CVPR,Nauta2021, multimodal_prototypical_nets} because of their interpretability and robustness properties \citep{sim_DNN_2022,yang2018robust}.
While limited work has been done in the NLP domain, PBNs have recently found application in text classification tasks such as propaganda detection~\citep{prototex}, logical fallacy detection~\citep{sourati2023robust}, sentiment analysis~\citep{protocnn}, and few-shot relation extraction~\citep{meng2023rapl}. 
ProseNet \citep{Ming_2019}, a prototype-based text classifier, uses several criteria for constructing prototypes~\citep{sparse_prototypes_for_text_2020}, and a special optimization procedure for better interpretability. ProtoryNet \citep{hong2020interpretable} leverages RNN-extracted prototype trajectories and deploys a pruning procedure for prototypes, and ProtoTex \citep{prototex} uses negative prototypes for handling the absence of features for classification. While PBNs are expected to be robust to perturbations, this property has not been systematically studied in NLP. Our paper consolidates PBN components used in prior studies and explores their robustness in different adversarial settings.

% Conclusions
\vspace{-0.6em}
\section{Conclusions}
\label{sec:conclusion}
\vspace{-0.6em}
Inspired by the state-of-the-art LMs and LLMs' lack of robustness to noisy data, we study the robustness of PBNs, as an architecturally robust variation of LMs, against both targeted and static adversarial attacks. We find that PBNs are more robust than vanilla LMs and even LLMs such as Llama3, both under targeted and static adversarial attack settings. Our results suggest that this robustness can be sensitive to hyperparameters involved in PBNs' training. More particularly, we note that a low number of prototypes and tight clustering conditions limit the robustness capacities of PBNs. Additionally, benefiting from the inherently interpretable architecture of PBNs, we showcase how learned prototypes can be utilized for robustness and also for gaining insights about their behavior facing adversarial perturbations, even when PBNs are wrong. In summary, our work provides encouraging results for the potential of PBNs to enhance the robustness of LMs across a variety of text classification tasks and quantifies the impact of architectural components on PBN robustness.

\section*{Acknowledgments}
This research was supported, in part, by the Army Research Laboratory under contract
W911NF-23-2-0183, and by the National Science Foundation under Contract No. IIS-2153546.

\section*{Limitations}
Although we cover a wide range of adversarial perturbations and strategies for their generation, we acknowledge that more complicated perturbations can also be created that are more effective and help the community have a more complete understanding of the models' robustness. Hence, we do not comment on the generalizability of our study to all possible textual perturbations besides our evaluation on AdvGLUE. Moreover, although it is customary in the field to utilize prototype-based networks for classification tasks, their application and robustness on other tasks remain to be explored. Furthermore, while we attempt to use a wide variety of backbones for our study, we do not ascertain similar patterns for all possible PBN backbones and leave this study for future work. Finally, we encourage more exploration of the interpretability of these models under different attacks to better understand the interpretability benefits of models when analyzing robustness.

\section*{Ethical Considerations}
Although the datasets and domains we focus on do not pose any societal harm, the potential harm that is associated with using the publicly available tools we used in this study to manipulate models in other critical domains should be considered. Issues surrounding anonymization and offensive content hold importance in data-driven studies, particularly in fields like natural language processing. Since we utilize datasets like IMDB, AG\_News, DBPedia, and AdvGLUE that are already content-moderated, there is no need for anonymization of data before testing for robustness in this study. 

\bibliography{acl_latex}

%\clearpage

\appendix

\section{Dataset Details}
\label{appendix:dataset-details}

The statistics of the datasets we used in this study to test the robustness of PBNs against perturbations are demonstrated in \autoref{tab:dataset-stats}. We present both statistics about the original dataset and statistics and details about the number of perturbations that we have gathered on each dataset with different attack strategies. All the original datasets we use in this study are gathered by other researchers and have been made public by them, mentioning non-commercial use, which aligns with how we use these datasets. We have included information on their descriptions and how they were gathered:

\begin{table*}[ht]
\centering
\small

\begin{tabular*}{1\textwidth}{@{\extracolsep{\fill}}cccccccccccc}
%\hline
\toprule
\textbf{Dataset} & \textbf{\#Classes} & \textbf{\#Tokens} &  \textbf{\#Train} & \textbf{\#Val} & \textbf{\#Test} & \textbf{BAE} & \textbf{DWB} & \textbf{PWWS} & \textbf{TB} & \textbf{TF} & \textbf{Other} \\ \midrule
%\cline{1-5}
IMDB  & 2 & 234 & 22,500 & 2,500 & 25,000  & 1784 &	1584 & 2816 & 2408 & 2880 & -\\
AG\_News & 4 & 103 & 112,400 & 7,600 & 7,600 & 663 & 1287 & 1533 & 1383 & 1893 & -\\
DBPedia & 9 & 38 & 240,942 & 36,003 & 60,794 & 1041 & 1143 & 1401 & 1281 & 1836 & -\\
\hline
SST\nobreakdashes-2 & 2 & 14 & 67,349 & 872 & 1,821 & - & - & - & - & - & 148\\
\bottomrule
\end{tabular*}
\caption{Dataset statistics: number of classes, the average number of tokens, and size of the perturbed datasets under BAE, DeepWordBug (DWB), PWWS, TextBugger (TB), TextFooler (TF), obtained. SST-2 subset comes from the AdvGlue benchmark~\citep{adv-glue} after removing the human-generated instances that do not belong to either category of perturbation classes.
}
\label{tab:dataset-stats}
\end{table*}

\paragraph{IMDB.} This dataset is compiled from a set of 50000 reviews sourced from IMDB in English, limiting each movie to a maximum of 30 reviews. It has maintained an equal count of positive and negative reviews, ensuring a 50\% accuracy through random guessing. To align with prior research on polarity classification, the authors specifically focus on highly polarized reviews. A review is considered negative if it scores $\leq$ 4 out of 10 and positive if it scores $\geq$ 7 out of 10. Neutral reviews are excluded from this dataset. Authors have made the dataset publicly available, and you can find more information about this dataset at \url{https://ai.stanford.edu/~amaas/data/sentiment/}.

\paragraph{AG\_News.} This dataset comprises over 1 million English news articles sourced from 2000+ news outlets over a span of more than a year by ComeToMyHead, an academic news search engine operational since July 2004. Provided by the academic community, this dataset aids research in data mining, information retrieval, data compression, data streaming, and non-commercial activities. This news topic classification dataset features four classes: world, sports, business, and science. The details about the intended use and access conditions are provided at \url{http://www.di.unipi.it/~gulli/AG_corpus_of_news_articles.html}.

\paragraph{DBPedia.} DBPedia\footnote{\url{https://www.dbpedia.org/}} seeks to extract organized information from Wikipedia's vast content. The gathered subset of data we used offers hierarchical categories for 342782 Wikipedia articles. These classes are distributed across three levels, comprising 9, 70, and 219 classes, respectively. We used the version that has nine classes: Agent, Work, Place, Species, UnitOfWork, Event, SportsSeason, Device, and TopicalConcept. Although the articles are in English, specific names (e.g., the name of a place or person) can be non-English. Find more information about this dataset at \url{https://huggingface.co/datasets/DeveloperOats/DBPedia_Classes}.

\paragraph{AdvGLUE.} Adversarial GLUE (AdvGLUE) \citep{adv-glue} introduces a multi-task English benchmark designed to investigate and assess the vulnerabilities of modern large-scale language models against various adversarial attacks. It encompasses five corpora, including SST-2 sentiment classification, QQP paraphrase test dataset, and QNLI, RTE, and MNLI, all of which are natural language inference datasets. To assess robustness, perturbations are applied to these datasets through both automated and human-evaluated methods, spanning word-level, sentence-level, and human-crafted examples. Our focus primarily centers on SST-2 due to its alignment with the other covered datasets in our study and its classification nature. This dataset has been made public by the authors and is released with CC BY-SA 4.0 license.

\section{Implementation Details}
\label{appendix:implementation-details}

\subsection{Experimental Environment}
For all the experiments that involved training or evaluating PBNs or vanilla LMs, we used three GPU NVIDIA RTX A5000 devices with Python v3.9.16 and CUDA v11.6, and each experiment took between 10 minutes to 2 hours, depending on the dataset and model used. All Transformer models were trained using the Transformers package v4.30.2 and Torch package v2.0.1+cu117. We used TextAttack v0.3.10 \citep{morris2020textattack} for implementing the employed attack strategies and perturbations.

\subsection{Training Details}
% \paragraph{Implementation details.}
% \label{subsec:PBN-components}
All prototypes are initialized randomly for a fair comparison, and only the last layer of LM backbones are trainable. The prototypes are trained without being constrained to a certain class from the beginning, and their corresponding class can be identified after training. The transformation from distances to class logits is done through a simple fully connected layer without intercept to avoid introducing additional complexity and keep the prediction interpretable through prototype distances. Both the backbone of PBNs and their vanilla counterparts leveraged the same LM and were fine-tuned separately to show the difference that is only attributed to the models' architecture. Focusing on the BERT-based PBN for evaluation, since BERT-base is one of the models from which we extract static perturbations by directly attacking it, to ensure generalization of the experiments on different backbones in the evaluation step, we use BERT-Medium \citep{turc2019wellread} as the backbone for BERT-based PBN and its vanilla counterpart.

For all the datasets, the training split, validation split, and test splits were used from \url{https://huggingface.co/}. During training on the IMDB, SST-2, and DBPedia datasets, the batch size was set to 64. This number was 256 on the AG\_News dataset. All the models were trained with the number of epochs adjusted according to an early stopping module with patience of 4 and a threshold value of 0.01 for change in accuracy.

All the Transformer models were fine-tuned on top of a pre-trained model gathered from \url{https://huggingface.co/}. Details of the models used in our experiments are presented in the following: 

\begin{itemize}
    \item Electra \citep{clark2020electra}: google/electra-base-discriminator;
    \item BART \citep{lewis2019bart}: ModelTC/bart-base-mnli;
    \item BERT \citep{devlin2018bert}: prajjwal1/bert-medium.
\end{itemize}

Furthermore, the models that were used in the process of gathering static perturbations were also pre-trained Transformer models gathered from \url{https://huggingface.co/}. Find the details of models used categorized by the dataset below:

\begin{itemize}
    \item IMDB: textattack/bert-base-uncased-imdb,  textattack/distilbert-base-uncased-imdb, textattack/roberta-base-imdb;
    \item AG\_News: textattack/bert-base-uncased-ag-news,  andi611/distilbert-base-uncased-ner-agnews, textattack/roberta-base-ag-news;
    \item DBPedia: dbpedia\_bert-base-uncased, dbpedia\_distilbert-base-uncased,  dbpedia\_roberta-base.
\end{itemize}

Since we could not find models from TextAttack \citep{morris2020textattack} library that were fine-tuned on DBPedia, the models that are presented above were fine-tuned by us on that dataset as well and then used as the target model.

\paragraph{Overhead of PBNs compared to vanilla LMs.} Since both PBNs and non-PBNs are trained similarly and the primary difference in their architecture is the involvement of prototypes in PBN architecture, the number of parameters in PBNs is only a fixed amount more than their non-PBN counterparts, based on the number of prototypes. This matter is depicted in \autoref{tab:overhead-analysis}.

\begin{table*}[h!]
\centering
\resizebox{\linewidth}{!}{%
\begin{tabular}{|c|c|c|c|}
\hline
Model Type  & BERT & ELECTRA & BART \\
\hline
Number of parameters & 110M & 110M & 139M \\
\hline
\# parameters more in PBNs (4 prototypes) & 1572864 (1.42\% more) & 1572864 (1.42\% more) & 1572864 (1.13\% more) \\
\hline
\# parameters more in PBNs (16 prototypes) & 6291456 (5.72\% more) & 6291456 (5.72\% more) & 6291456 (4.53\% more) \\
\hline
\# parameters more in PBNs (64 prototypes) & 25165824 (22.87\% more) & 25165824 (22.87\% more) & 25165824 (18.10\% more) \\
\hline
\end{tabular}}
\caption{Comparison of model parameters between PBNs and non-PBNs based on the number of prototypes.}
\label{tab:overhead-analysis}
\end{table*}

Furthermore, the overhead of using prototypes in the inference time is close to zero and negligible. However, because of the prototypes in the PBN architecture, their training, and satisfying the extra objective functions, training PBNs will take longer than non-PBN models, which similarly depends on the number of prototypes (e.g., up to a 5\% increase compared to non-PBNs using 16 prototypes). 

\subsection{GPT4o and Llama3 Baseline} 
We used GPT4o \citep{gpt_4o_2024} and Llama3 \citep{llama3modelcard} as baselines in our experiments to compare their performance on original and perturbed examples with PBNs and their vanilla counterparts. In this section, we present the prompts that we gave to these models to generate the baseline responses and the reported performance in \autoref{tab:new-table-results-both}. We used the following prompts for the four different datasets:

IMDB: \noindent\textit{Identify the binary sentiment of the following text:} \textbf{[text]}. \textit{Strictly output only "negative" or "positive" according to the sentiment and nothing else. Assistant:}

AG\_News: \noindent\textit{Categorize the following news strictly into only one of the following classes: world, sports, business, and science. Ensure that you output only the category name and nothing else. Text:} \textbf{[text]}. \textit{Assistant:}

DBPedia: \noindent\textit{Categorize the following text article strictly into only one taxonomic category from the following list: Agent, Work, Place, Species, UnitOfWork, Event, SportsSeason, Device, and TopicalConcept. Ensure that you output only the category name and nothing else. Text:} \textbf{[text]}. \textit{Assistant:}

SST-2: \noindent\textit{Identify the binary sentiment of the following text:} \textbf{[text]}. \textit{Strictly output only "negative" or "positive" according to the sentiment and nothing else. Assistant:}

\paragraph{Analysis of LLMs' predictions.} One potential reason for the underperformance of LLMs on certain tasks could be limitations in the evaluation framework. In other words, while LLMs may generate correct predictions, the evaluation method might fail to recognize or appropriately assess them. To address this, we conducted a sanity check to determine whether the LLMs’ predictions were within the label distributions of the datasets. Across all experiments, only 0.5\% of the predicted labels fell outside the dataset’s label distribution. Additionally, when we adjusted the prompt to ask the model for indices corresponding to correct labels using a label-to-index dictionary, we observed similar patterns and results.

\subsection{Perturbation Details}
\label{subsec:perturbation-details}

\paragraph{BAE.} BAE (BERT-based Adversarial Examples; \citealp{garg-ramakrishnan-2020-bae}) generates adversarial examples for text classification by leveraging the BERT masked language model (MLM) to create contextually appropriate token replacements and insertions. BAE replaces and inserts tokens in the original text by masking a portion of the text and leveraging the BERT-MLM to generate alternatives for the masked tokens. This approach ensures the adversarial examples maintain grammaticality and semantic coherence better than prior methods, leading to more effective and natural-looking adversarial attacks. BAE has been shown to significantly reduce the accuracy of even robust classifiers by employing these perturbations.

\paragraph{TextFooler.} TextFooler \citep{jin2020bert} is an adversarial attack method designed to generate adversarial text examples that can fool natural language processing models while maintaining semantic similarity and grammatical correctness. The approach operates in a black-box setting, where the attacker has no knowledge of the model's architecture or parameters. TextFooler works by first identifying the most important words in the target text that influence the model's prediction. It then iteratively replaces these words with their most semantically similar and grammatically correct synonyms until the model's prediction changes. This method ensures that the adversarial examples remain human-readable and convey the same meaning as the original text, thus preserving utility while effectively deceiving the model.

\paragraph{TextBugger.} TextBugger \citep{li2018textbugger} is an attack framework designed to generate adversarial texts that deceive deep learning-based text understanding (DLTU) systems while maintaining readability and semantic coherence for human readers. It operates under both white-box and black-box settings. In the white-box scenario, TextBugger identifies critical words by analyzing the model's gradients, then applies one of five perturbation techniques—such as inserting spaces, deleting characters, swapping adjacent characters, or substituting with similar words or characters—to create adversarial examples. In the black-box scenario, it uses sentence importance and word scoring to select target words for manipulation. These perturbations are crafted to be subtle yet effective in misleading text classifiers used for tasks like sentiment analysis and toxic content detection, achieving high success rates while preserving the original text's utility for humans.

\paragraph{PWWS.} The Probability Weighted Word Saliency (PWWS; \citealp{ren-etal-2019-generating}) algorithm is designed to generate adversarial examples for text classification by substituting words with synonyms. Words are prioritized for a synonym-swap transformation based on a combination of their saliency score and maximum word-swap effectiveness. This approach ensures that the adversarial examples are lexically and grammatically correct while maintaining semantic similarity to the original text, making them difficult for humans to detect.

\paragraph{DeepWordBug.} DeepWordBug \citep{gao2018black} is a method designed to generate small perturbations in text that lead to misclassifications by deep-learning classifiers, even in a black-box setting. It utilizes unique scoring strategies to identify key tokens within the text that, when altered, can cause incorrect predictions. The approach applies simple character-level transformations to these critical tokens, ensuring minimal changes to the original text while still altering the classification.

\section{Additional Experiments}

\begin{table}
\aboverulesep=0ex % Solution part 1 of 3
\belowrulesep=0ex
\centering
% \small
\setlength\tabcolsep{4.5pt}
\resizebox{\linewidth}{!}{%
\begin{tabular}{@{\extracolsep{\fill}}rr|r|r|r}
\toprule
& \multicolumn{1}{c|}{AG\_News} & \multicolumn{1}{c|}{DBPedia} & \multicolumn{1}{c}{IMDB} & \multicolumn{1}{c}{SST2}\\
\midrule
BART & \textbf{93.8} & \underline{91.4} & \underline{97.5} & \underline{93.1} \\
+ PBN & 93.5 & \textbf{92.2} & 97.3 & 90.0 \\
\midrule
BERT & 92.6 & 90.9 & 95.6 & 83.9 \\
+ PBN & 92.9 & 90.4 & 95.3 & 77.8 \\
\midrule
ELEC. & 93.1 & 90.6 & 96.1 & 87.6 \\
+ PBN & \underline{93.6} & 90.9 & 95.9 & \textbf{98.5} \\
\midrule
GPT4o & 71.4 & 68.4 & \textbf{99.4} & 91.0\\
\midrule
Llama3 & 68.2 & 49.8 & 93.6 & 76.0 \\
\bottomrule
\end{tabular}}
\caption{Comparison between PBNs, vanilla LMs, GPT4o, and Llama3 on the original datasets. The best performance for each dataset among all models is \textbf{boldfaced}, and the second best performance is \underline{underlined}.}
\label{tab:performance-of-models-on-original-datasets}
\end{table}

\subsection{Performance of PBNs on Original Datasets}
\label{app:performance-of-pbns-on-original-datasets}

The performance of PBN models compared with both non-PBN models, GPT4o, and Llama 3, are shown in \autoref{tab:performance-of-models-on-original-datasets}. The results suggest that smaller fine-tuned language models perform better than LLMs (i.e., GPT4o and Llama3) on original datasets, and PBNs and non-PBNs perform on par.

\begin{table}
\aboverulesep=0ex % Solution part 1 of 3
\belowrulesep=0ex
\centering
% \small
\setlength\tabcolsep{4.5pt}
\resizebox{\linewidth}{!}{%
\begin{tabular}{@{\extracolsep{\fill}}rrr|rr|rr}
\toprule
& \multicolumn{2}{c|}{AG\_News} & \multicolumn{2}{c|}{DBPedia} & \multicolumn{2}{c}{IMDB} \\
 & Orig & Adv & Orig & Adv & Orig & Adv \\
\midrule
BART & 93.7 & 92.6 & 91.2 & 91.3 & 97.5 & 96.0 \\
+ PBN & 93.2 & 93.8 & 92.0 & 91.6 & 97.2 & 97.0 \\
\midrule
BERT & 92.5 & 91.0 & 90.8 & 90.5 & 95.5 & 94.2 \\
+ PBN & 92.8 & 91.2 & 90.3 & 90.8 & 95.2 & 95.0 \\
\midrule
ELEC. & 93.0 & 92.1 & 90.5 & 90.0 & 96.0 & 94.5 \\
+ PBN & 93.5 & 91.8 & 90.8 & 89.7 & 95.8 & 95.0 \\
\bottomrule
\end{tabular}}
\caption{Comparison between PBNs and vanilla LMs on the original and paraphrased version of texts from AG\_News, DBPedia, and IMDB datasets that GPT3.5 generated.}
\label{tab:robustness-of-pbns-paraphrased-based-perturbations}
\end{table}

\subsection{Robustness of PBNs Against Paraphrased-Based Perturbations}
\label{app:subsec:robustness-pbns-paraphrased-based-perturbations}

Comparison between PBNs and vanilla LMs on the original and paraphrased version of texts from AG\_News, DBPedia, and IMDB datasets that GPT3.5 generated are shown in \autoref{tab:robustness-of-pbns-paraphrased-based-perturbations}, which illustrated that both PBNs and vanilla LMs are robust to such perturbations.

\begin{table*}
\aboverulesep=0ex % Solution part 1 of 3
\belowrulesep=0ex
\centering
\small
\setlength\tabcolsep{4.5pt}
\resizebox{\linewidth}{!}{%
\begin{tabular}{@{\extracolsep{\fill}}rrrrrr|rrrrr|rrrrr}
\multicolumn{6}{c}{\normalsize Using the best hyperparameters}  \\
\toprule
& \multicolumn{5}{c|}{AG\_News} & \multicolumn{5}{c|}{DBPedia} & \multicolumn{5}{c}{IMDB} \\
 & BAE & DWB & PWWS & TB & TF & BAE & DWB & PWWS & TB & TF & BAE & DWB & PWWS & TB & TF \\
% Model &  &  &  &  &  &  &  &  &  &  &  &  &  &  &  \\
\midrule
BART & 8.7 & \textbf{26.9} & 20.8 & 35.7 & 25.0 & 9.1 & \textbf{27.3} &\textbf{ 16.9} & \textbf{50.1} & \textbf{26.2} & 4.1 & 6.4 & 4.2 & 33.3 & 5.9 \\
+ PBN & \textbf{9.0} & 24.8 & \textbf{22.2} & \textbf{37.7} & \textbf{27.6} & \textbf{10.1} & 17.1 & 15.9 & 43.3 & 26.0 & \textbf{4.7} & \textbf{6.6} & \textbf{8.1} & \textbf{33.4} & \textbf{13.6} \\
\midrule
BERT & 7.4 & \textbf{26.8} & 21.6 & 37.4 & 24.1 & 9.7 & 27.9 & 19.4 & \textbf{53.8} & 28.8 & 4.0 & 5.7 & 4.4 & 30.1 & 5.0 \\
+ PBN & \textbf{7.7} & 26.6 & \textbf{24.1} &\textbf{ 37.7} & \textbf{28.8 }& \textbf{10.9} & \textbf{27.9} & \textbf{22.4} & 50.0 & \textbf{30.6} & \textbf{4.6} & \textbf{6.7} & \textbf{9.3} & \textbf{35.9} & \textbf{15.4} \\
\midrule
ELEC. & \textbf{8.2} & \textbf{23.7} & 17.5 & \textbf{32.7} & 20.8 & 10.9 & 24.6 & 17.7 & \textbf{58.0} & 22.9 & 5.4 & 8.1 & 8.8 & \textbf{44.7} & 11.2 \\
+ PBN & 8.1 & 21.2 & \textbf{18.9} & 31.8 & \textbf{24.0} & \textbf{11.9} & \textbf{25.1} & \textbf{19.4} & 48.5 & \textbf{26.8} & \textbf{5.6} & \textbf{8.4} & \textbf{13.3} & 38.6 & \textbf{18.5} \\
\bottomrule
\end{tabular}}
\newline
\vspace{0.3em}
\resizebox{\linewidth}{!}{%
\begin{tabular}{@{\extracolsep{\fill}}rrrrrr|rrrrr|rrrrr}
\multicolumn{7}{c}{\large Averaged over all hyperparameters}  \\
\toprule
& \multicolumn{5}{c|}{AG\_News} & \multicolumn{5}{c|}{DBPedia} & \multicolumn{5}{c}{IMDB} \\
 & BAE & DWB & PWWS & TB & TF & BAE & DWB & PWWS & TB & TF & BAE & DWB & PWWS & TB & TF \\
\midrule
BART & \textbf{8.7} & \textbf{26.9} & \textbf{20.8} & \textbf{35.7} & 25.0 & 9.1 & \textbf{27.3} & \textbf{16.9} & \textbf{50.1} & \textbf{26.2} & 4.1 & \textbf{6.4} & 4.2 & \textbf{33.3} & 5.9 \\
+ PBN & 8.3 & 19.3 & 20.5 & 32.6 & \textbf{25.2} & \textbf{9.7} & 17.1 & 15.9 & 40.4 & 24.7 & \textbf{4.4} & 6.1 & \textbf{6.5} & 29.5 &\textbf{ 10.1} \\
\midrule
BERT & \textbf{7.4} & \textbf{26.8} & 21.6 & \textbf{37.4} & 24.1 & \textbf{9.7} & \textbf{27.9} & \textbf{19.4} & \textbf{53.8} & \textbf{28.8} & 4.0 & \textbf{5.7} & 4.4 & \textbf{30.1} & 5.0 \\
+ PBN & 7.2 & 24.1 & \textbf{21.9} & 35.0 & \textbf{25.9} & 9.5 & 24.1 & 19.3 & 43.1 & 27.6 & \textbf{4.1} & 5.5 & \textbf{5.0} & 27.3 & \textbf{7.1} \\
\midrule
ELEC. & \textbf{8.2} & \textbf{23.7} &\textbf{ 17.5} & \textbf{32.7} & \textbf{20.8} & \textbf{10.9} & \textbf{24.6} & \textbf{17.7} & \textbf{58.0} & 22.9 & 5.4 & \textbf{8.1} & 8.8 & \textbf{44.7} & 11.2 \\
+ PBN & 7.7 & 15.3 & 16.1 & 26.1 & 20.1 & 10.2 & 18.2 & 16.6 & 40.2 & \textbf{23.7} & 5.4 & 6.7 & \textbf{10.1} & 31.3 & \textbf{13.6} \\
\bottomrule
\end{tabular}}
\caption{
Comparison of PBNs and vanilla LMs' robustness with respect to Average Percentage of Words Perturbed (APWP) under targeted adversarial attack perturbations, both using the best hyperparameters and averaged over all hyperparameters for PBNs, on IMBD, AG\_News, and DBPeida datasets, under BAE, DeepWordBug (DWB), PWWS, TextBugger (TB), TextFooler (TF). The highest APWP showing the superior model for each architecture is \textbf{boldfaced}.}
\label{tab:robustness-of-pbns-apwp}
\end{table*}

\subsection{Robustness of PBNs' w.r.t. Average Percentage of Words Perturbed}
\label{app:subsec:robustness-pbns-using-best-hp-apwp}

The Comparison of PBNs and vanilla LMs' robustness with respect to the Average Percentage of Words Perturbed (APWP) under different adversarial settings, different datasets, and perturbation strategies is shown in \autoref{tab:robustness-of-pbns-apwp}. We observed that while using the best hyperparameters, PBNs are more robust than vanilla LMs in the majority of the cases, this superiority is less salient when averaging over all hyperparameters involved in PBNs' training, which entails how PBNs' robustness is sensitive to hyperparameters. 

\begin{table*}
\aboverulesep=0ex % Solution part 1 of 3
\belowrulesep=0ex
\centering
\small
\setlength\tabcolsep{4.5pt}
\resizebox{\linewidth}{!}{%
\begin{tabular}{@{\extracolsep{\fill}}rrrrrr|rrrrr|rrrrr}
\multicolumn{12}{c}{\normalsize Targeted Attacks; Attack Success Rate (ASR \%) reported}  \\
\toprule
& \multicolumn{5}{c|}{AG\_News} & \multicolumn{5}{c|}{DBPedia} & \multicolumn{5}{c}{IMDB} \\
 & BAE & DWB & PWWS & TB & TF & BAE & DWB & PWWS & TB & TF & BAE & DWB & PWWS & TB & TF \\
% Model &  &  &  &  &  &  &  &  &  &  &  &  &  &  &  \\
\midrule
BART & 14.8 & 53.2 & 53.6 & 31.8 & 76.5 & 18.9 & 28.3 & 43.1 & 21.1 & 71.9 & 74.1 & 74.7 & 99.3 & 78.5 & 100.0 \\
+ PBN & \textbf{14.8} & \textbf{40.4} & \textbf{50.7} & \textbf{29.8} & \textbf{76.2} & \textbf{17.0} & \textbf{14.7} & \textbf{28.7} & \textbf{12.7} & \textbf{49.4} & \textbf{55.5} & \textbf{49.2} & \textbf{86.2} & \textbf{49.7} & \textbf{88.5} \\
\midrule
BERT & 17.0 & 78.0 & 69.8 & 45.7 & 88.8 & 13.9 & 24.8 & 31.6 & 22.0 & 61.3 & 82.5 & 79.7 & 99.9 & 83.9 & 99.9 \\
+ PBN & \textbf{14.0} & \textbf{64.7} & \textbf{57.0} & \textbf{39.3} & \textbf{82.1} & \textbf{13.5} & \textbf{23.4} & \textbf{27.6} & \textbf{19.6} & \textbf{51.3} & \textbf{68.4} & \textbf{61.8} & \textbf{91.3} & \textbf{74.0} & \textbf{92.4} \\
\midrule
ELEC. & 24.8 & 89.5 & 69.1 & 87.8 & 87.9 & 14.5 & 42.8 & 45.6 & 42.3 & 75.3 & 52.5 & 49.2 & 95.3 & 67.8 & 99.3 \\
+ PBN & \textbf{18.5} & \textbf{50.4} & \textbf{55.7} & \textbf{63.6} & \textbf{80.0} & \textbf{12.6} & \textbf{19.4} & \textbf{26.1} & \textbf{27.1} & \textbf{46.5} & \textbf{41.0} & \textbf{35.9} & \textbf{77.7} & \textbf{55.6} & \textbf{86.2} \\
\bottomrule
\end{tabular}}
\newline
\vspace{0.3em}
\resizebox{\linewidth}{!}{%
\begin{tabular}{@{\extracolsep{\fill}}rrrrrr|rrrrr|rrrrr|r}
\multicolumn{9}{c}{\large Static Attacks; Accuracy (\%) reported}  \\
\toprule
& \multicolumn{5}{c|}{AG\_News} & \multicolumn{5}{c|}{DBPedia} & \multicolumn{5}{c|}{IMDB} & SST2\\
 & BAE & DWB & PWWS & TB & TF & BAE & DWB & PWWS & TB & TF & BAE & DWB & PWWS & TB & TF & GLUE\\
\midrule
BART & \underline{53.2} & \underline{76.7} & \underline{83.2} & \underline{77.5} & \underline{85.8} & 55.5 & \underline{68.6} & 58.4 & \underline{72.5} & \underline{71.3} & \underline{74.1} & \underline{80.5} & \underline{83.6} & \underline{85.8} & \underline{87.6} & \underline{29.8} \\
+ PBN & 50.4 & 68.3 & 75.7 & 70.5 & 79.6 & \underline{56.4} & 65.8 & \underline{58.7} & 70.9 & 69.5 & 69.2 & 78.7 & 79.7 & 81.9 & 78.3 & \textbf{37.6} \\
+ Aug. & \textbf{71.7} & \textbf{78.4} & \textbf{85.5} & \textbf{77.6} & \textbf{90.1} & \textbf{84.0} & \textbf{79.6} & \textbf{89.7} & \textbf{88.8} & \textbf{94.0} & \textbf{85.7} & \textbf{86.7} & \textbf{92.9} & \textbf{89.9} & \textbf{96.5} & - \\
\midrule
BERT & 47.8 & 64.0 & 75.9 & 69.4 & 80.7 & 62.3 & \underline{61.4} & \underline{75.4} & \underline{78.4} & \underline{82.0} & \underline{75.1} & \underline{77.1} & \underline{85.0} & \underline{83.4} & \underline{85.9} & \underline{42.0} \\
+ PBN & \underline{49.5} & \underline{66.2} & \underline{76.4} & \textbf{71.3} & \underline{82.3} & \underline{63.5} & 61.1 & 73.9 & 76.9 & 79.4 & 71.0 & 73.9 & 81.1 & 80.2 & 79.2 & \textbf{47.1} \\
+ Aug. & \textbf{58.3} & \textbf{71.6} & \textbf{78.3} & \underline{71.2} & \textbf{85.4} & \textbf{75.5} & \textbf{70.9} & \textbf{84.1} & \textbf{90.5} & \textbf{91.0} & \textbf{83.2} & \textbf{77.6} & \textbf{91.7} & \textbf{90.8} & \textbf{89.2} & - \\
\midrule
ELECTRA & 50.4 & \textbf{65.0} & \underline{73.5} & \underline{63.9} & 77.8 & \underline{79.7} & \underline{66.9} & \underline{80.9} & \underline{81.4} & \underline{84.4} & \textbf{89.7} & \underline{90.3} & \underline{94.6} & \underline{94.5} & \underline{95.6} & \underline{44.3} \\
+ PBN & \underline{52.7} & \underline{63.9} & \textbf{73.7} & \textbf{67.1} & 77.8 & 73.4 & 64.1 & 73.0 & 76.4 & 80.6 & 80.6 & 79.4 & 79.9 & 80.2 & 86.8 & \textbf{56.4} \\
+ Aug. & \textbf{55.0} & 59.5 & 71.7 & 61.6 & \textbf{79.5} & \textbf{86.2} & \textbf{73.8} & \textbf{88.1} & \textbf{84.5} & \textbf{92.8} & \underline{89.4} & \textbf{93.7} & \textbf{95.3} & \textbf{94.9} & \textbf{95.8} & - \\
\bottomrule
\end{tabular}}
\caption{
Comparison of PBNs and vanilla LMs (+ vanilla LMs with adversarial augmented training under static attack setting) under both targeted and static adversarial attack perturbations, averaged over all hyperparameters for PBNs, on IMBD, AG\_News, DBPeida (+ SST-2 AdvGLUE under static attack setting) datasets, under BAE, DeepWordBug (DWB), PWWS, TextBugger (TB), TextFooler (TF). The highest accuracy and lowest ASR showing the superior model for each architecture is \textbf{boldfaced}, and the second best model is \underline{underlined} for static attacks.}
\label{tab:robustness-of-pbn-on-average}
\end{table*}

\subsection{Robustness of PBNs' Averaged over Hyperparameters}
\label{app:subsec:robustness-pbns-averaged-over-hps}

The comparison of PBNs and vanilla LMs under different adversarial settings, on different datasets, and different attacking strategies, averaged over all hyperparameters of PBNs, is shown in \autoref{tab:robustness-of-pbn-on-average}. Comparing the observed trends with the trends observed when using the best hyperparameters for PBNs, our results suggested that PBNs' robustness is sensitive to hyperparameters that are involved in their training. 

\subsection{Effect of Distance Function on Robustness}
\label{app:subsec:effect-of-distance-function}

\begin{figure}
    \centering
    \includegraphics[width=1.0\columnwidth]{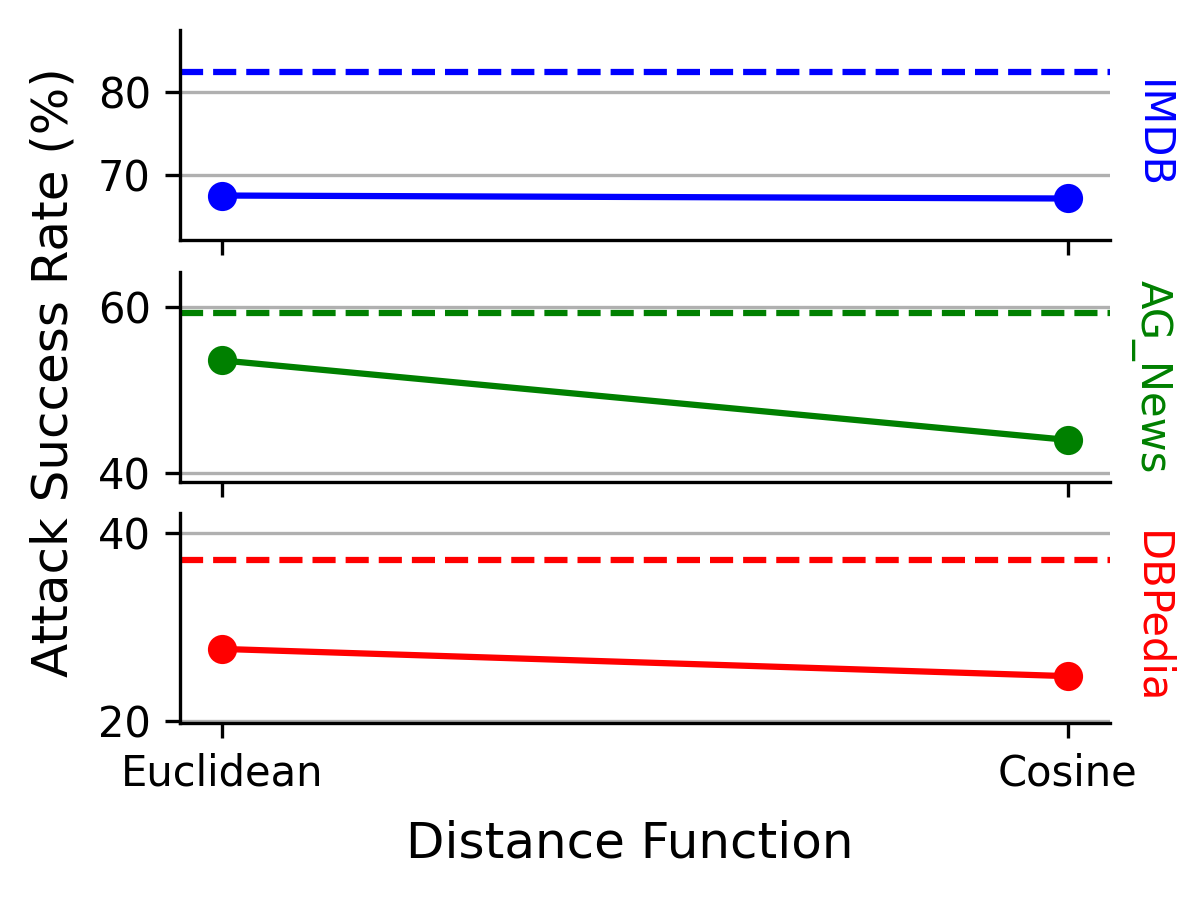}
    \caption{Attack Success Rate (ASR \%) of PBNs with different distance functions and other hyperparameters set to their best values and averaged across other possible variables (e.g., backbone and attack type). The dashed line represents the ASR for the vanilla LMs.}
    \label{fig:asr_effect_of_distance_func}
    \vspace{-0.3cm}
\end{figure}

\begin{figure}
    \centering
    \includegraphics[width=1.0\columnwidth]{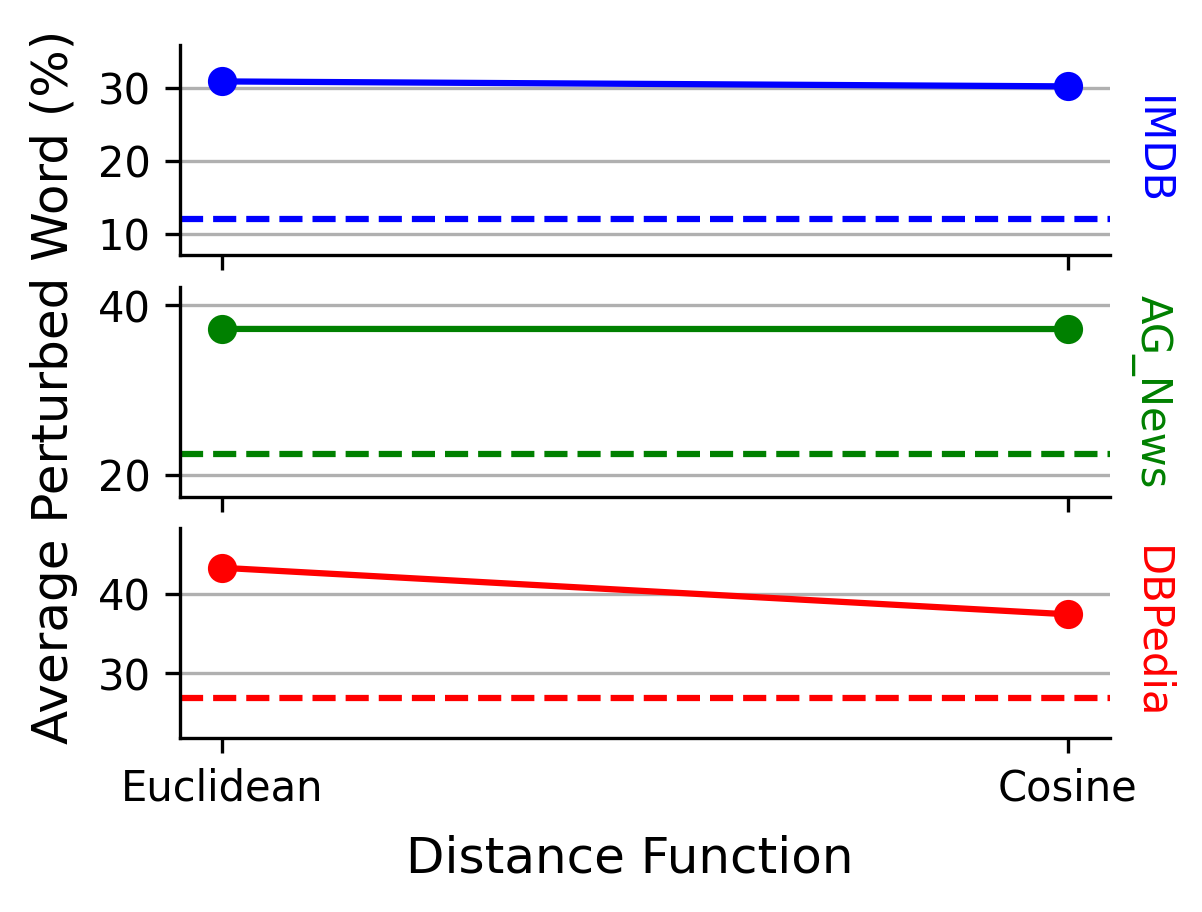}
    \caption{Average Percentage of Words Perturbed (APWP) of PBNs with different distance functions and other hyperparameters set to their best values and averaged across other possible variables (e.g., backbone and attack type). The dashed line represents the APWP for the vanilla LMs.}
    \label{fig:apwp_effect_of_distance_func}
    \vspace{-0.3cm}
\end{figure}

\begin{figure}
    \centering
    \includegraphics[width=1.0\columnwidth]{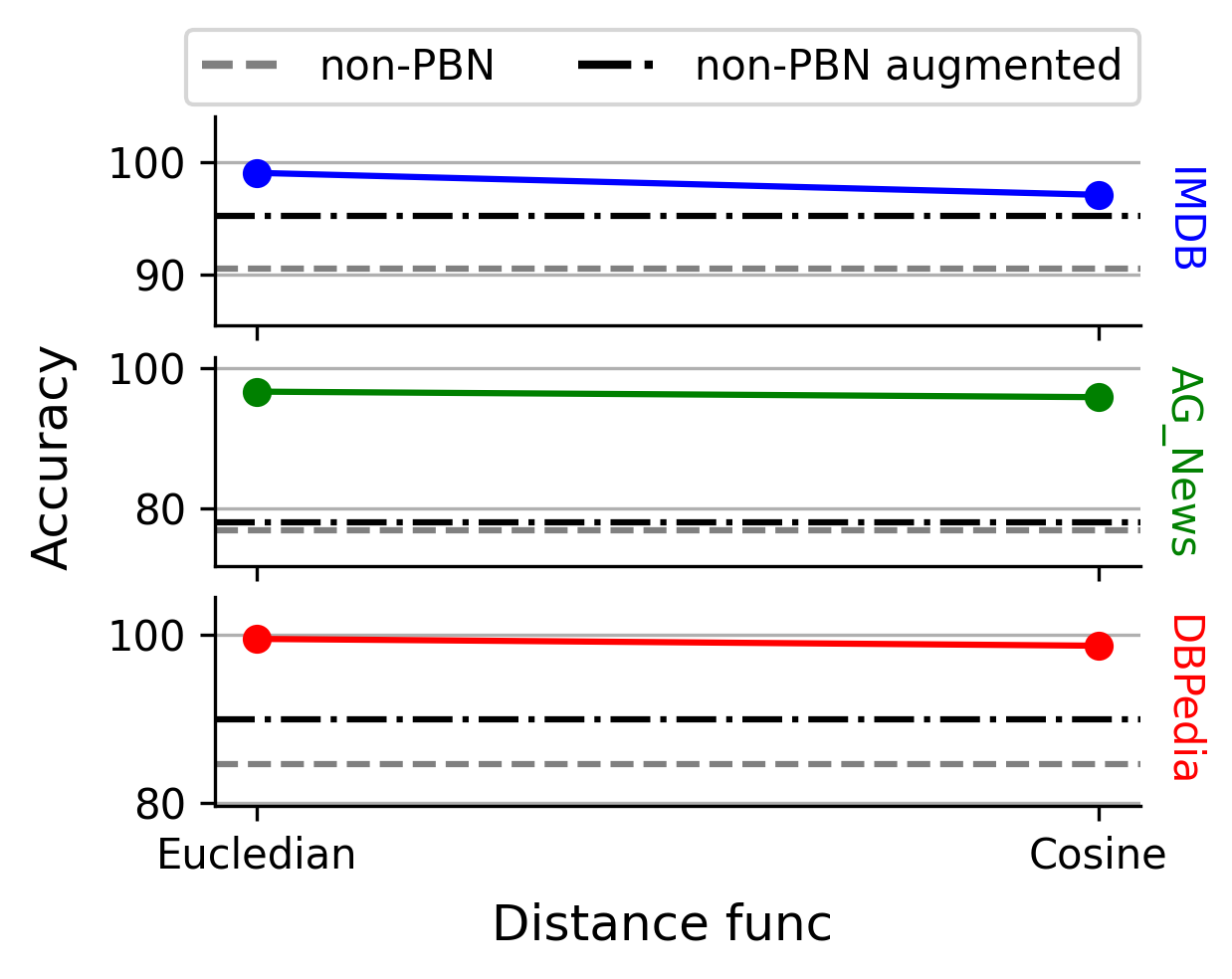}
    \caption{Accuracy of PBNs under static adversarial settings, with different distance functions, with other hyperparameters set to their best values and averaged across other possible variables (e.g., backbone and attack type). The dashed line represents the accuracy for the vanilla LMs.}
    \label{fig:acc_effect_of_distance_func}
    \vspace{-0.3cm}
\end{figure}

\autoref{fig:asr_effect_of_distance_func}, \autoref{fig:apwp_effect_of_distance_func}, and \autoref{fig:acc_effect_of_distance_func} illustrate the robustness of PBNs compared to vanilla LMs, using different distance functions, showing that PBNs' robustness is not sensitive to this hyperparameter. 

\subsection{Effect of Interpretability on Robustness}
\label{app:subsec:effect-of-interpretability-robustneess}

\begin{figure}
    \centering
    \includegraphics[width=1.0\columnwidth]{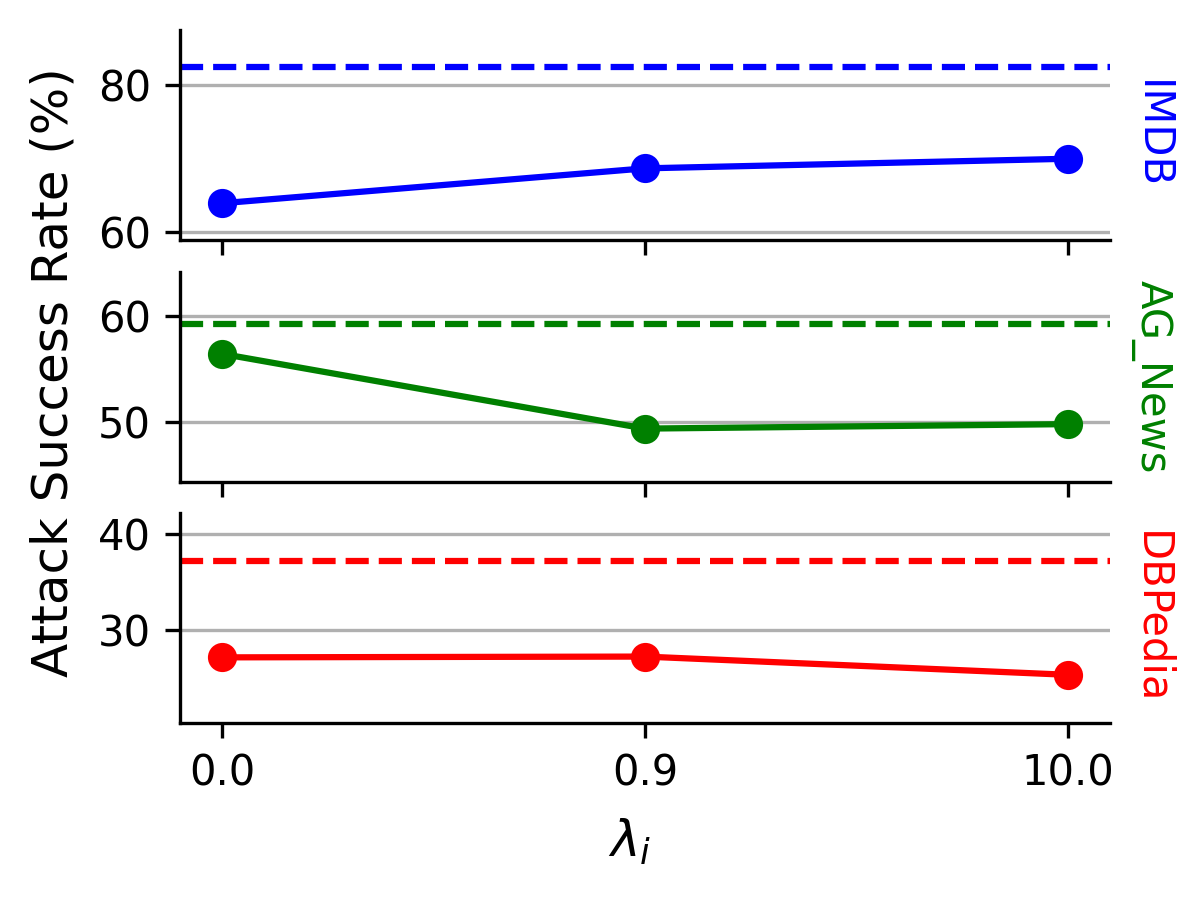}
    \caption{Attack Success Rate (ASR \%) of PBNs with different $\lambda_i$ values adjusting the importance of interpretability of the prototypes in training, with other hyperparameters set to their best values, and averaged across other possible variables (e.g., backbone and attack type). The dashed line represents the ASR for the non-PBN model.}
    \label{fig:asr_effect_of_pi}
\end{figure}

\begin{figure}
    \centering
    \includegraphics[width=1.0\columnwidth]{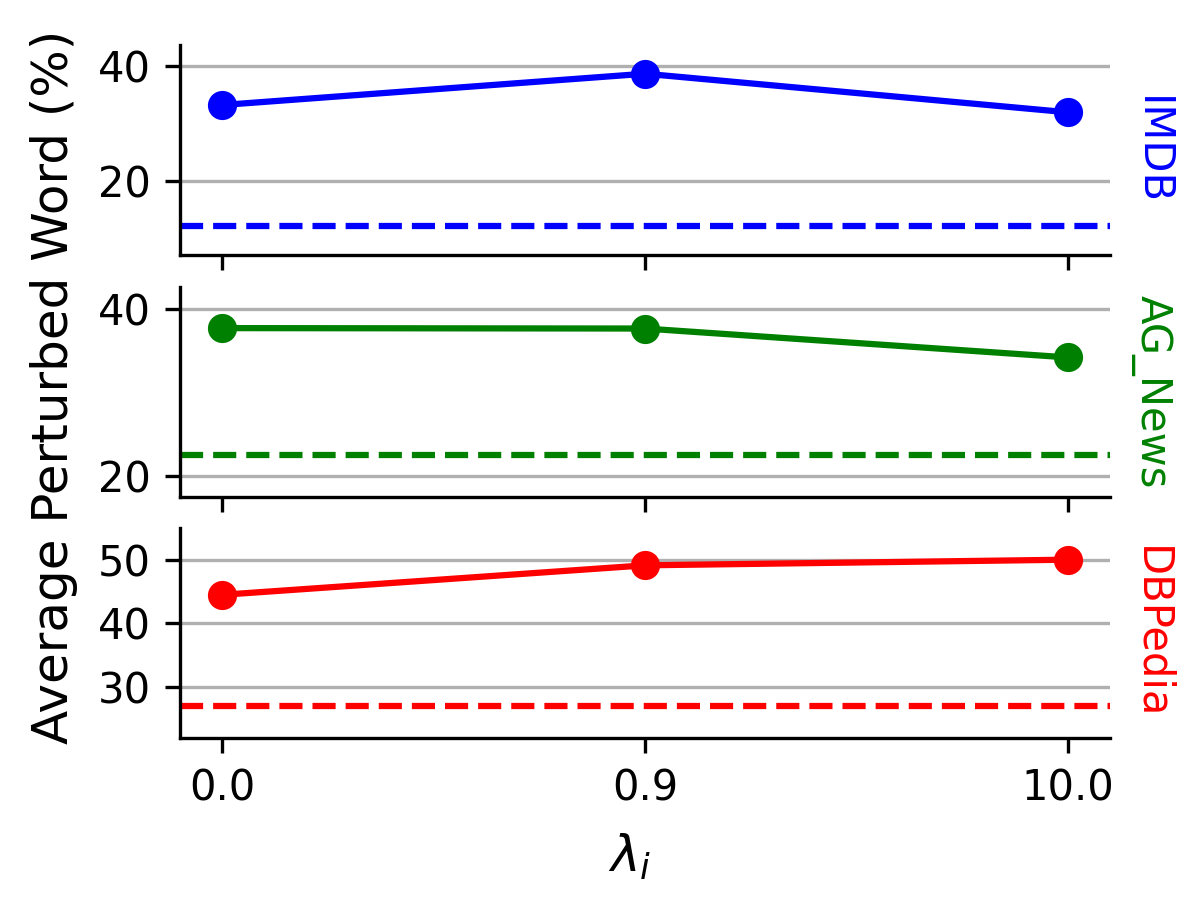}
    \caption{Average Percentage of Words Perturbed (APWP) of PBNs with different $\lambda_i$ values adjusting the importance of interpretability of the prototypes in training, with other hyperparameters set to their best values, and averaged across other possible variables (e.g., backbone and attack type). The dashed line represents the APWP for the non-PBN model.}
    \label{fig:apwp_effect_of_pi}
\end{figure}

\begin{figure}
    \centering
    \includegraphics[width=1.0\columnwidth]{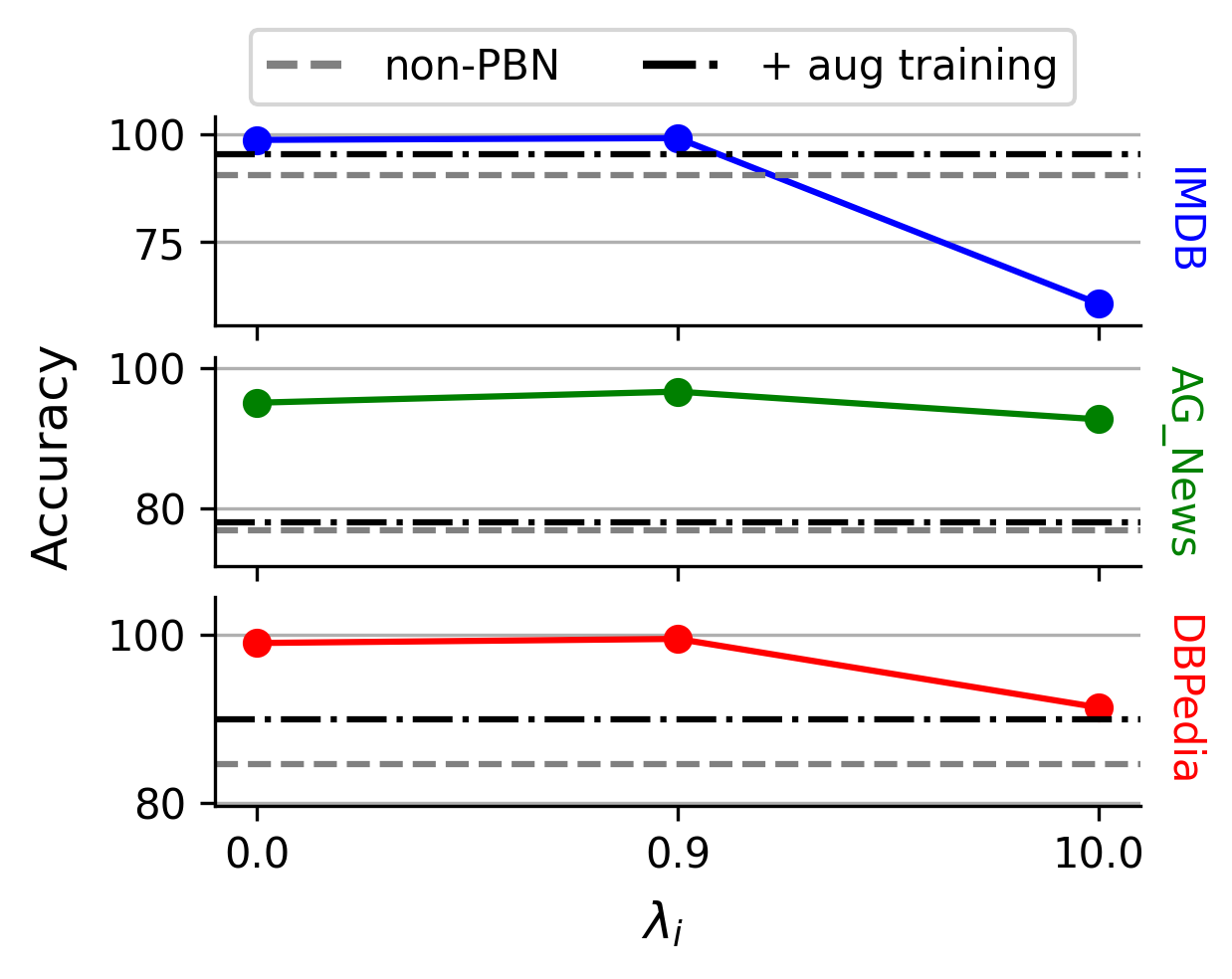}
    \caption{Accuracy of PBNs under static adversarial settings, with different $\lambda_i$ values adjusting the level of interpretability in PBNs, with other hyperparameters set to their best values and averaged across other possible variables (e.g., backbone and attack type). The dashed line represents the accuracy for the vanilla LMs.}
    \label{fig:acc_effect_of_pi}
    \vspace{-0.3cm}
\end{figure}

\autoref{fig:asr_effect_of_pi}, \autoref{fig:apwp_effect_of_pi}, and \autoref{fig:acc_effect_of_pi} illustrate the robustness of PBNs compared to vanilla LMs, using different values of $\lambda_i$ adjusting the importance of interpretability, showing that overall, PBNs' robustness is not sensitive to this hyperparameter. 

\subsection{Effect of Clustering on Robustness}
\label{app:subsec:effect-of-clustering-robustneess}

\begin{figure}
    \centering
    \includegraphics[width=1.0\columnwidth]{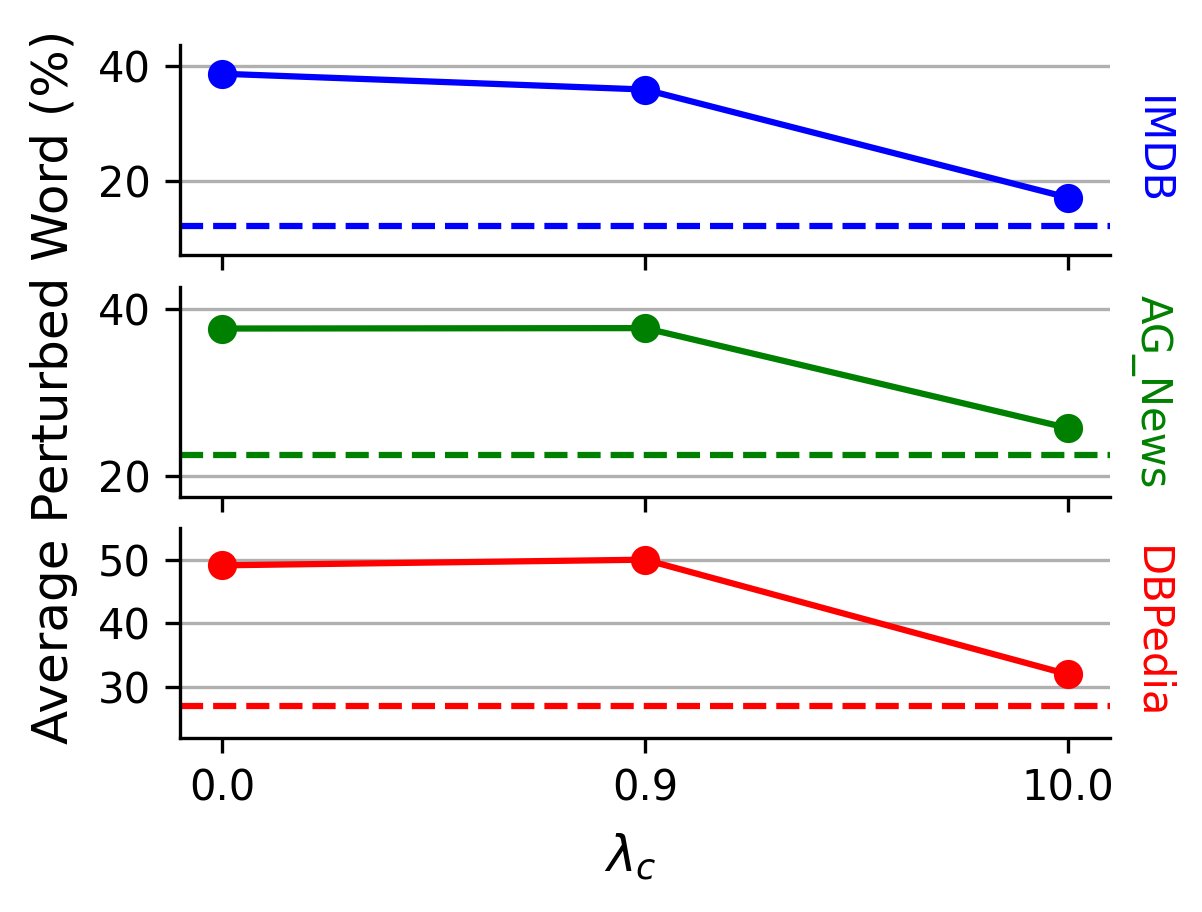}
    \caption{Average Percentage of Words Perturbed (APWP) of PBNs with different $\lambda_c$ values adjusting the importance of clustering of examples in PBNs, with other hyperparameters set to their best values, and averaged across other possible variables (e.g., backbone and attack type). The dashed line represents the APWP for the non-PBN model.}
    \label{fig:apwp_effect_of_pc}
\end{figure}

\begin{figure}
    \centering
    \includegraphics[width=1.0\columnwidth]{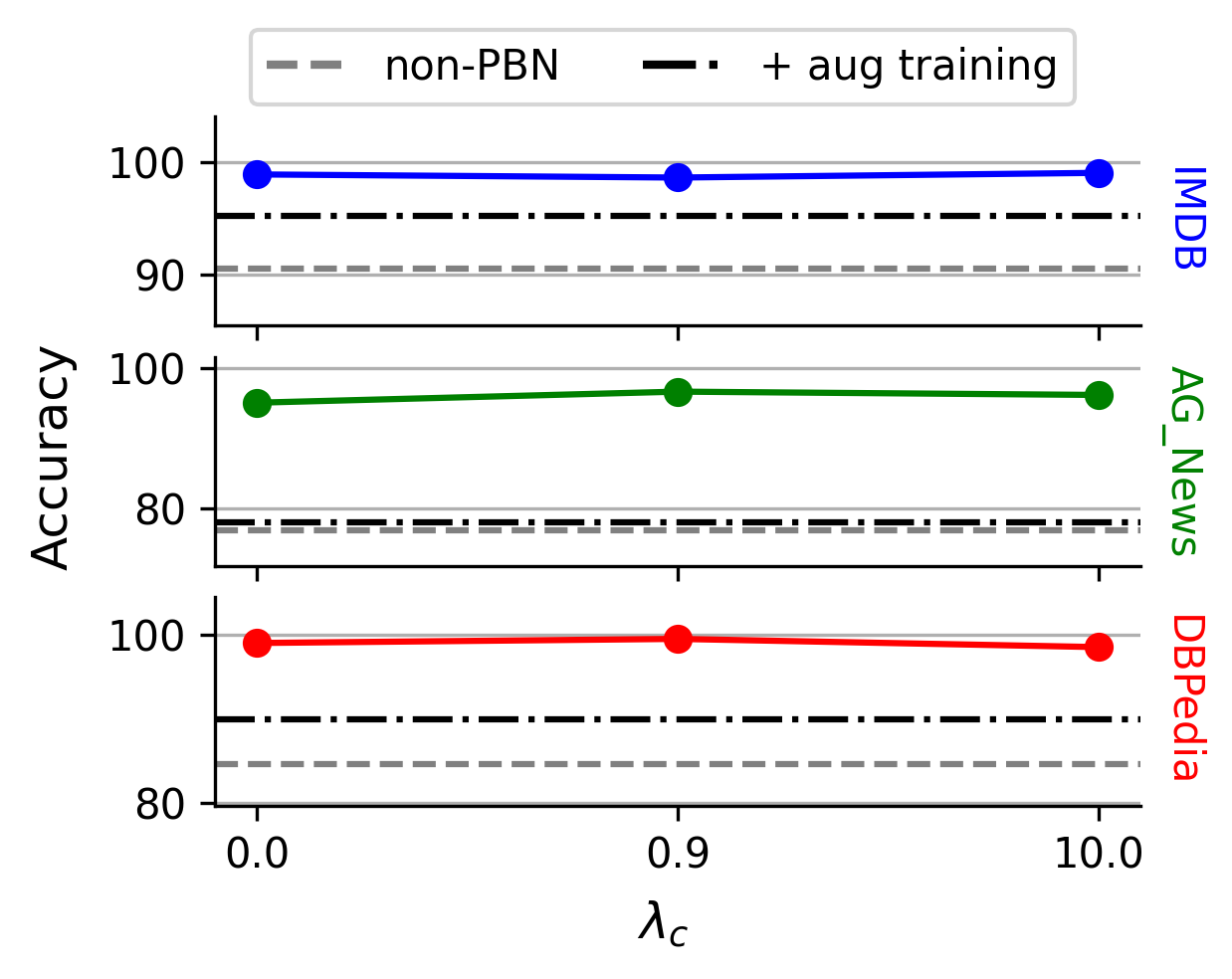}
    \caption{Accuracy of PBNs under static adversarial settings, with different $\lambda_c$ values adjusting the level of clustering in PBNs, with other hyperparameters set to their best values and averaged across other possible variables (e.g., backbone and attack type). The dashed line represents the accuracy for the vanilla LMs.}
    \label{fig:acc_effect_of_pc}
    \vspace{-0.3cm}
\end{figure}

\autoref{fig:apwp_effect_of_pc}, \autoref{fig:acc_effect_of_pc} illustrate the robustness of PBNs compared to vanilla LMs, using different values of $\lambda_c$ adjusting the importance of clustering, that alongside the trends observed using ASR (see \autoref{fig:asr_effect_of_pc}), show that overall, PBNs' robustness degrades with tighter clustering in PBNs' training. 

\subsection{Effect of Separation on Robustness}
\label{app:subsec:effect-of-separation-robustneess}

\begin{figure}
    \centering
    \includegraphics[width=1.0\columnwidth]{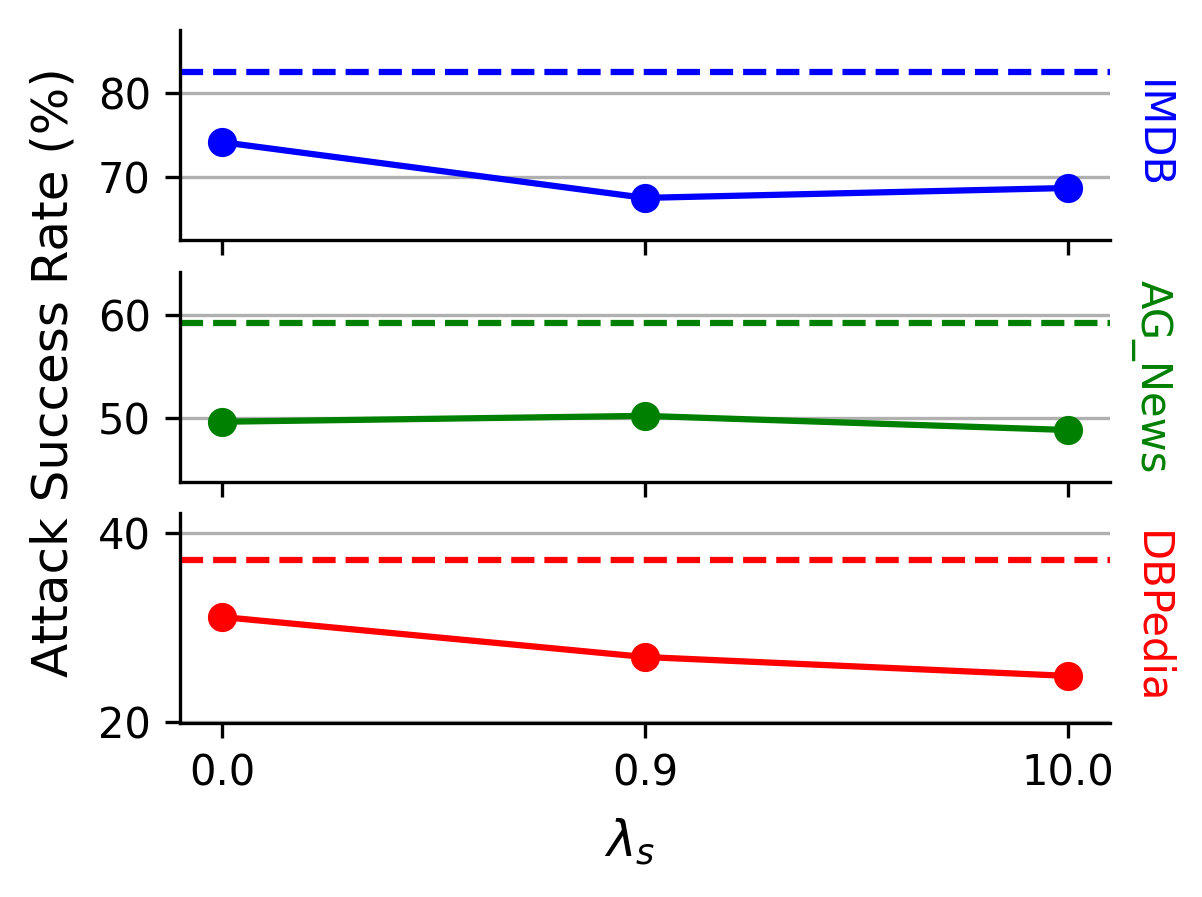}
    \caption{Attack Success Rate (ASR \%) of PBNs with different $\lambda_s$ values adjusting the level of separation between the prototypes, with other hyperparameters set to their best values and averaged across other possible variables (e.g., backbone and attack type). The dashed line represents the ASR for the vanilla LMs.}
    \label{fig:asr_effect_of_ps}
    \vspace{-0.3cm}
\end{figure}

\begin{figure}
    \centering
    \includegraphics[width=1.0\columnwidth]{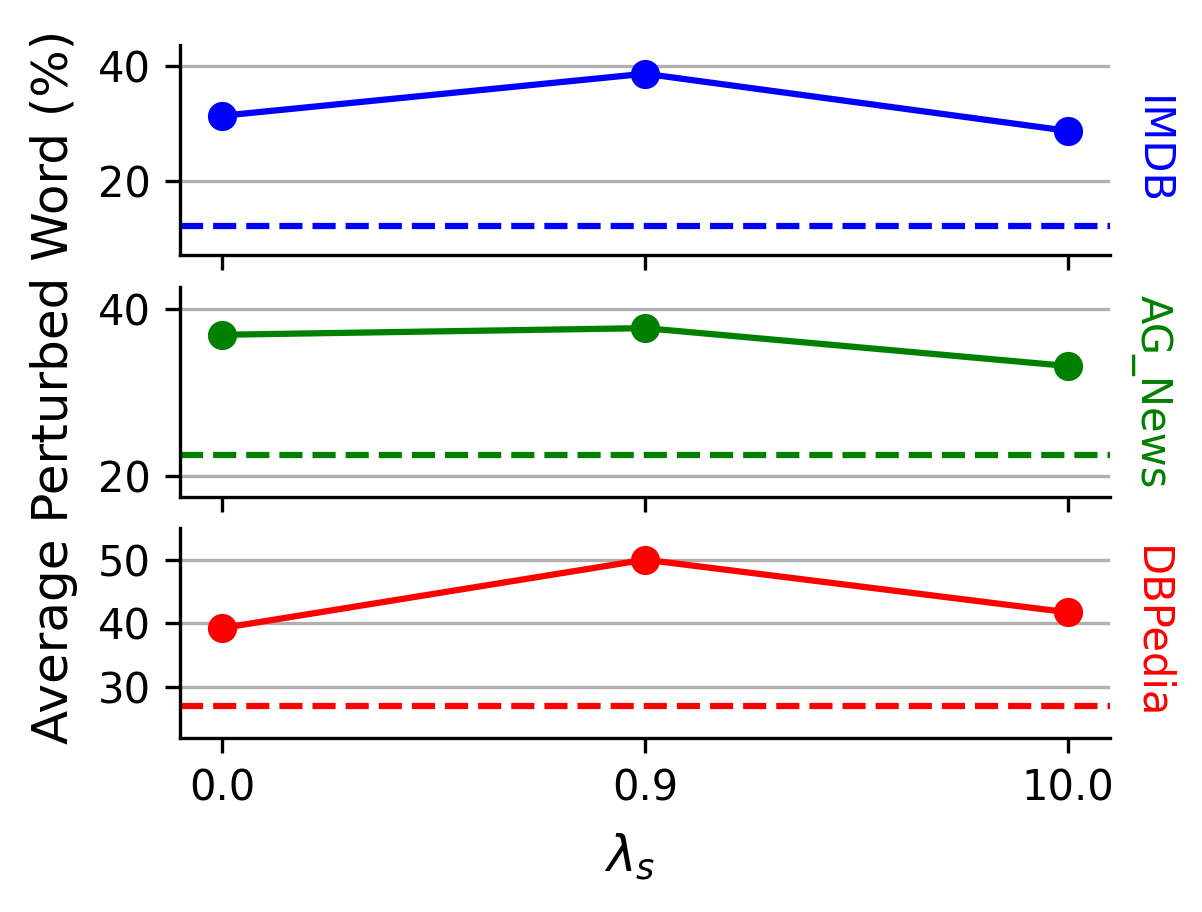}
    \caption{Average Percentage of Words Perturbed (APWP) of PBNs with different $\lambda_s$ values adjusting the level of separation between the prototypes, with other hyperparameters set to their best values and averaged across other possible variables (e.g., backbone and attack type). The dashed line represents the APWP for the vanilla LMs.}
    \label{fig:apwp_effect_of_ps}
    \vspace{-0.3cm}
\end{figure}

\begin{figure}
    \centering
    \includegraphics[width=1.0\columnwidth]{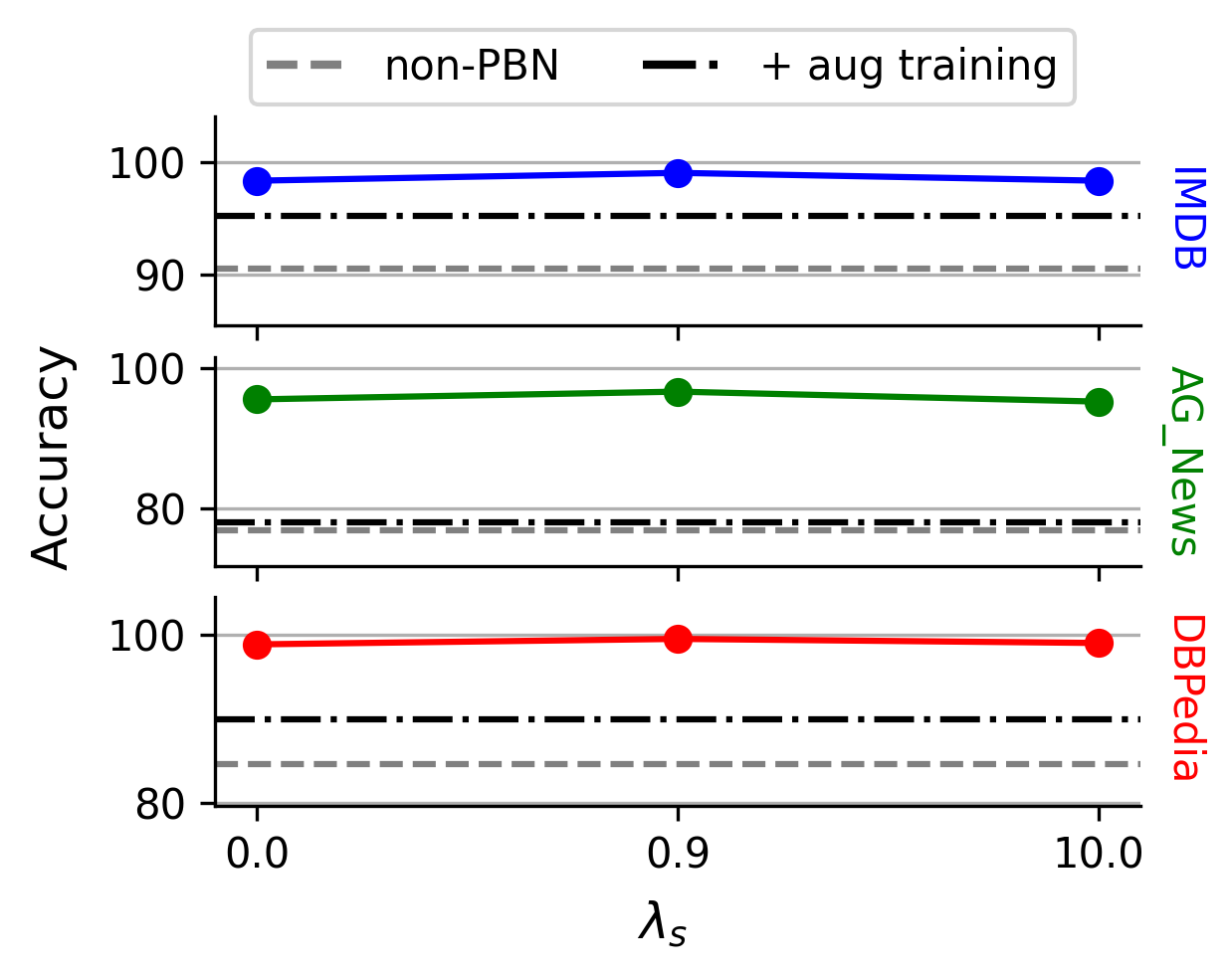}
    \caption{Accuracy of PBNs under static adversarial settings, with different $\lambda_s$ values adjusting the level of separation between the prototypes, with other hyperparameters set to their best values and averaged across other possible variables (e.g., backbone and attack type). The dashed line represents the accuracy for the vanilla LMs.}
    \label{fig:acc_effect_of_ps}
    \vspace{-0.3cm}
\end{figure}

\autoref{fig:asr_effect_of_ps}, \autoref{fig:apwp_effect_of_ps}, and \autoref{fig:acc_effect_of_ps} illustrate the robustness of PBNs compared to vanilla LMs, using different values of $\lambda_s$ adjusting the importance of separability between prototypes, showing that overall, PBNs' robustness is not sensitive to this hyperparameter. 

\subsection{Effect of Number of Prototypes on Robustness}
\label{app:subsec:effect-of-num-prototypes-on-robustness}

\begin{figure}
    \centering
    \includegraphics[width=1.0\columnwidth]{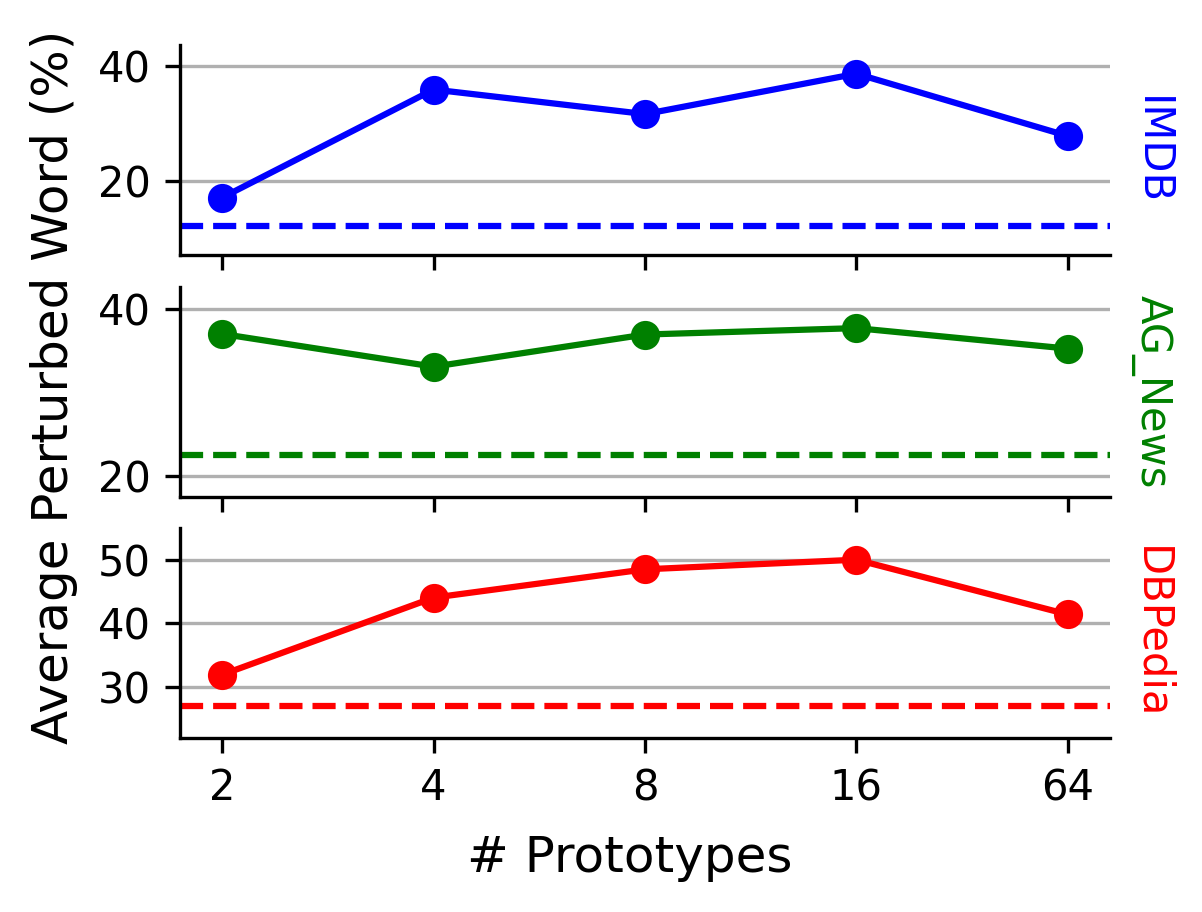}
    \caption{Average Percentage of Words Perturbed (APWP) of PBNs with different numbers of prototypes, with other hyperparameters set to their best values, and averaged across other possible variables (e.g., backbone and attack type). The dashed line represents the APWP for the non-PBN model.}
    \label{fig:apwp_effect_of_num_protos}
\end{figure}

\begin{figure}
    \centering
    \includegraphics[width=1.0\columnwidth]{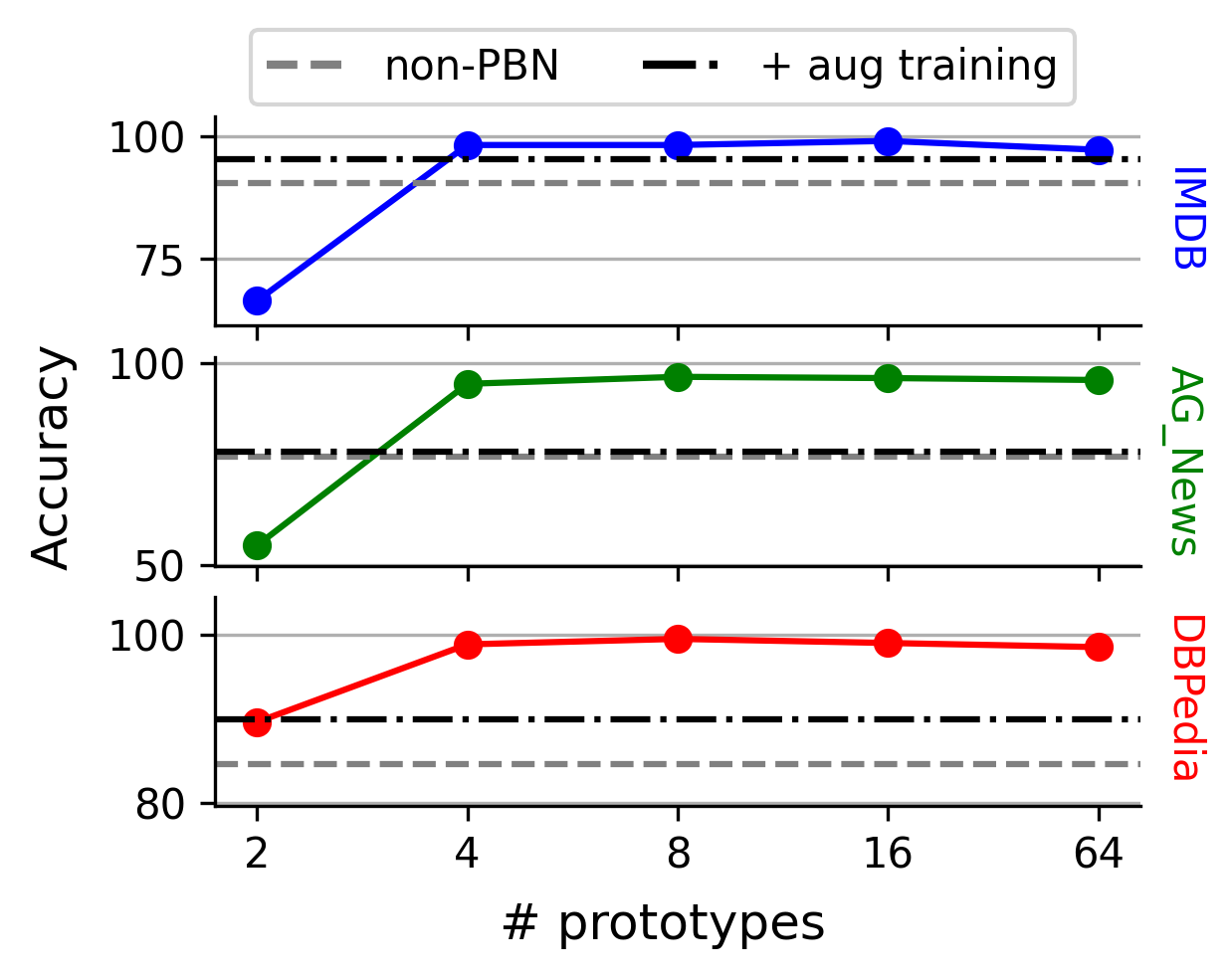}
    \caption{Accuracy of PBNs under static adversarial settings, with different numbers of prototypes, with other hyperparameters set to their best values and averaged across other possible variables (e.g., backbone and attack type). The dashed line represents the accuracy for the vanilla LMs.}
    \label{fig:acc_effect_of_num_protos}
    \vspace{-0.3cm}
\end{figure}

\autoref{fig:apwp_effect_of_num_protos}, \autoref{fig:acc_effect_of_num_protos} illustrate the robustness of PBNs compared to vanilla LMs, using different numbers of prototypes, that alongside the trends observed using ASR (see \autoref{fig:effect-of-num-prototypes}), show that overall, PBNs' robustness degrades with low number of prototypes as PBNs can capture lower number of semantic patterns in such conditions, resulting in lower robustness.

\end{document}